\documentclass[12pt]{article}

\usepackage{a4wide}
\usepackage{titlesec}
\usepackage[center]{caption}
\usepackage{amsmath}
\usepackage{authblk}
\usepackage{setspace}
\usepackage{floatrow}
\usepackage{listings}
\usepackage{subcaption}
\usepackage[hyphens]{url}
\usepackage{natbib}
\usepackage{rotating}
\usepackage{float}
\usepackage{threeparttable}
\usepackage{array}
\usepackage{multirow}
\usepackage{appendix}
\usepackage{todonotes} 
\usepackage[multiple]{footmisc}
\usepackage[utf8]{inputenc} 
\usepackage[T1]{fontenc}    
\usepackage{xcolor}
\renewcommand{\footnotesize}{\scriptsize}
\definecolor{perplexityteal}{HTML}{20808D}
\usepackage[colorlinks=true, 
            linkcolor=perplexityteal, 
            citecolor=perplexityteal, 
            urlcolor=perplexityteal]{hyperref}
\usepackage{url}            
\usepackage{booktabs}       
\usepackage{nicefrac}       
\usepackage{lipsum}
\usepackage{graphicx}       
\usepackage{newtxtext}
\usepackage{hyperref} 

\newcounter{en}
\title{\normalfont The Adoption and Usage of AI Agents: \\
Early Evidence from Perplexity\thanks{J.Y. and N.Y. contributed equally. We thank Gustav Lindqvist, Alexis Weill, and many other Perplexity staff for helpful insights, discussions, and technical assistance. All errors are the authors' own. Correspondence to \href{mailto:jeryang@hbs.edu}{\texttt{jeryang@hbs.edu}} and \href{mailto:jerry@perplexity.ai}{\texttt{jerry@perplexity.ai}}.}
}

\makeatletter
\def\@maketitle{%
  \newpage
  \begin{center}%
    {\LARGE \@title \par}%
    \vskip 1.5em%
    {
      Jeremy Yang\textsuperscript{1} \ Noah Yonack\textsuperscript{2} \\[0.3ex] 
      Kate Zyskowski\textsuperscript{2} \ Denis Yarats\textsuperscript{2} \ Johnny Ho\textsuperscript{2} \ Jerry Ma\textsuperscript{2}\\[1em]
      \textsuperscript{1}Harvard University \quad \\ 
      \vskip 0.5em
      \textsuperscript{2}Perplexity \\
    }
    \vskip 1.5em
    December 7, 2025
  \end{center}%
  \par
  \vskip 1.5em}
\makeatother

\begin{document}
\maketitle

\begin{abstract}
This paper presents the first large-scale field study of the adoption, usage intensity, and use cases of general-purpose AI agents operating in open-world web environments. Our analysis centers on Comet, an AI-powered browser developed by Perplexity, and its integrated agent, Comet Assistant. Drawing on hundreds of millions of anonymized user interactions, we address three fundamental questions: Who is using AI agents? How intensively are they using them? And what are they using them for? Our findings reveal substantial heterogeneity in adoption and usage across user segments. Earlier adopters, users in countries with higher GDP per capita and educational attainment, and individuals working in digital or knowledge-intensive sectors—such as digital technology, academia, finance, marketing, and entrepreneurship—are more likely to adopt or actively use the agent. To systematically characterize the substance of agent usage, we introduce a hierarchical agentic taxonomy that organizes use cases across three levels: topic, subtopic, and task. 
The two largest topics—\textit{Productivity \& Workflow} and \textit{Learning \& Research}—account for 57\% of all agentic queries, while the two largest subtopics—\textit{Courses} and \textit{Shopping for Goods}—make up 22\%. The top 10 out of 90 tasks represent 55\% of queries. Personal use constitutes 55\% of queries, while professional and educational contexts comprise 30\% and 16\%, respectively. In the short term, use cases exhibit strong stickiness, but over time, users tend to shift toward more cognitively oriented topics. The diffusion of increasingly capable AI agents carries important implications for researchers, businesses, policymakers, and educators, inviting new lines of inquiry into this rapidly emerging class of AI capabilities.
\end{abstract}

\clearpage
\section{Introduction}

2025 is frequently heralded as the year of agentic AI, as the frontier shifts from conversational Large Language Model (LLM) chatbots to action-oriented AI agents.\footnote{\url{https://finance.yahoo.com/news/nvidia-jensen-huang-says-ai-044815659.html}\newline \hspace*{1.8em} \url{https://x.com/gdb/status/1879327050819104778}\newline \hspace*{1.8em} \url{https://www.aboutamazon.com/news/company-news/amazon-ceo-andy-jassy-on-generative-ai}\newline \hspace*{1.8em} \url{https://www.ibm.com/think/insights/ai-agents-2025-expectations-vs-reality}} This narrative has emerged as AI agents have progressed from a largely theoretical construct to widely productized assistants, demonstrating strong potential to transform work and daily life by planning and executing complex tasks in response to high-level human instructions with little supervision \citep{wooldridge1995intelligent}.\footnote{Examples of such agentic AI products or features include Perplexity's Comet browser; OpenAI's ChatGPT Operator, Codex, and Atlas browser; Anthropic's Claude Code and Computer Use; Google's Gemini Assistant; and Microsoft's Copilot.} AI agents could profoundly reshape individual workflows, as well as organizational and market structures, by increasing productivity and efficiency and lowering transaction costs as autonomous participants in both consumption and production processes \citep{hadfield2025economy, rothschild2025agentic, shahidi2025coasean}. In aggregate, Precedence Research estimates that the global agentic AI market size will grow from \$8 billion in 2025 to \$199 billion by 2034.\footnote{\url{https://www.precedenceresearch.com/agentic-ai-market}} PwC forecasts that the overall associated economic contribution could reach between \$2.6 trillion and \$4.4 trillion annually by 2030.\footnote{\url{https://www.pwc.com/m1/en/publications/agentic-ai-the-new-frontier-in-genai.html}} 

Despite this enthusiasm and its far-reaching economic implications, systematic behavioral evidence on how people actually adopt and use AI agents in the field remains limited, often relying on non-representative firm surveys \citep{pan2025measuring} or focusing on specialized agents such as coding assistants \citep{sarkar2025ai}.\footnote{\url{https://knowledge.wharton.upenn.edu/special-report/2025-ai-adoption-report/}} Launched in July 2025, Comet by Perplexity is among the first widely adopted AI browsers and offers the embedded Comet Assistant as a general-purpose AI agent capable of performing user-specified tasks across open-world web environments. By studying hundreds of millions of anonymized user interactions with Comet and Comet Assistant, we narrow the gap by providing early insights into three fundamental questions: Who is using AI agents? How intensively are they using them? And what are they using them for?

\subsection*{AI Agents}\label{definition} 
We define agentic AI systems as AI assistants capable of autonomously pursuing user-defined goals by planning and taking multi-step actions on a user's behalf to interact with and effect outcomes across real-world environments.

In general, agentic AI is a concept that resists precise definition. Despite variations, the definitions share several common themes: goal orientation, action taking, and autonomy. For instance, \cite{shavit2023practices} defines agentic AI systems as those capable of taking actions that consistently contribute toward achieving goals over extended periods without their behavior being explicitly specified in advance, and \cite{schluntz2024building} describes agents as systems that dynamically direct their own processes and tool use, maintaining control over how they complete tasks. \cite{pplx2025agent} refines these definitions by replacing the term ``agent'' with ``assistant,'' arguing that each AI agent is best understood as a personal, powerful generalist serving the interests of a single user or customer, in contrast to a human agent who typically manages multiple clients within narrow professional roles or licensing constraints and often faces conflicting incentives. In addition, we place particular emphasis on the agent’s ability not only to exchange information with its environment but also to actively modify it. 

Under the ReAct framework, an agentic workflow typically cycles automatically between three iterative phases to achieve the end goal: thinking, acting, and observing \citep{yao2022react}.\footnote{\url{https://huggingface.co/learn/agents-course/en/unit1/agent-steps-and-structure}} In the thinking phase, the agent interprets the goal from the query and devises a step-by-step plan to achieve it.\footnote{We use query and prompt interchangeably.} In the acting phase, the agent executes actions by controlling external tools to interact with its environment. In the observing phase, the agent processes feedback from its environment and returns to the thinking phase to confirm or revise its plan as needed.

It is also useful to contrast LLM chatbots and AI agents. Both chatbots and agents build on LLMs, but agents extend chatbots' capabilities beyond conversations to include autonomous actions. LLMs serve as the “brain” of an agent, functioning as the central reasoning engine that processes information, evaluates options, and makes decisions. Tools are the “hands” that connect the agent’s reasoning to the external world, enabling it to act upon its environment. More advanced agent capabilities also include multi-agent orchestration—the ability to interface with and manage workflows across multiple collaborating agents—and self-evolution—the ability to identify gaps in pre-specified resources and dynamically expand them.\footnote{\url{https://www.kaggle.com/whitepaper-introduction-to-agents}}

\subsection*{Research Setting: Perplexity and Comet}

Perplexity is an AI-powered platform that helps users discover, analyze, and act on information. Instead of requiring users to navigate through pages of results (``blue links''), as traditional search engines do, Perplexity interacts with the web on users' behalf to deliver direct, verifiable, and conversational answers. Each answer includes inline citations and links to original sources, enabling users to verify information and explore topics in more detail.\footnote{\url{https://www.perplexity.ai/help-center/en/articles/10352155-what-is-perplexity}} 

Comet is a browser from Perplexity that embeds an AI assistant directly into the browsing experience, helping users discover, analyze, and act on information more effectively. Its core feature, Comet Assistant, operates as an autonomous agent that takes actions and completes open-world web-based tasks on behalf of users. To fulfill user requests, Comet Assistant can execute a variety of tasks, including scheduling meetings, editing documents, sending emails, booking flights, making purchases, and more.\footnote{We provide some sample agentic queries in Figure \ref{fig:agent prompt} and an example of the agent executing a real task in Figure \ref{fig:agent flow} in Appendix \ref{agent demo}.} 

Comet was launched on July 9, 2025, on desktop for subscribers to Perplexity's Max tier\footnote{Perplexity offers three consumer subscription tiers: Free, Pro (\$20 per month), and Max (\$200 per month).}, along with selected users from a pre-launch waitlist.\footnote{\url{https://www.perplexity.ai/hub/blog/introducing-comet}} Access expanded to Pro subscribers on August 13, 2025, beginning with users in the United States.\footnote{\url{https://www.perplexity.ai/hub/blog/the-intelligent-business-introducing-comet-for-enterprise-pro}} On October 2, 2025, Comet became available to all users worldwide.\footnote{\url{https://www.perplexity.ai/hub/blog/comet-is-now-available-to-everyone-worldwide}} In addition to these general cohorts, Comet was opened to university students globally on September 3, 2025.

\subsection*{Data}

Our analysis relies on three samples collected from Comet desktop users between July 9 and October 22, 2025.\footnote{We define Comet users as those who made at least one query on Comet during our study period. We use October 22, 2025, as the cutoff date because a major agent update began rolling out to selected users on October 23, which could affect adoption and usage patterns thereafter. The new agent was launched to all users on November 6. The updated agent performs 23\% better than the previous version and offers greater multitasking capacity across multiple tabs. The agent we analyze in our data operates in a single web environment. For more details, see: \url{https://www.perplexity.ai/hub/blog/the-new-comet-assistant}.} First, we use anonymized data from the entire population of Comet users and their queries to provide high-level, aggregated statistics on agent adoption and usage intensity; this sample includes millions of users and hundreds of millions of queries. Second, we analyze a random sample of 100,000 Comet users and classify their O*NET occupation clusters and subclusters based on the National Career Clusters Framework to examine variation across occupations.\footnote{\url{https://www.onetonline.org/find/career?c=0}\newline \hspace*{1.8em} \url{https://careertech.org/career-clusters/}} Third, we analyze a separate random sample of 100,000 agent users and classify all of their agentic queries using a novel hierarchical agentic taxonomy to better understand common use cases at the topic, subtopic, and task levels.

\subsection*{Summary of Findings}


We report two sets of results on the adoption and use of AI agents: the extensive and intensive margins and a comprehensive taxonomy of use cases.

\subsubsection*{Adoption and usage intensity}
Overall, agent adoption and usage intensity demonstrate sustained growth with acceleration following the general availability (GA) of Comet. The post-GA period accounts for 60\% of agent adopters and 50\% of agentic queries throughout our sampling period. Earlier Comet adopters (those with pre-GA access) represent a disproportionately large share of agent adopters and agentic queries relative to their user share. The disparity is more pronounced in usage intensity than in adoption—an average user in the first cohort (July 9) is twice as likely to adopt the agent and makes nine times as many agentic queries as an average user in the GA cohort (October 2). At the country level, adoption and usage intensity show strong positive correlations with GDP per capita and average years of education. At the occupational level, adopters and queries tend to come more from digital or knowledge-intensive domains. Digital technology\footnote{The Digital Technology Career Cluster focuses on developing digital systems for communication and data storage using critical technologies such as artificial intelligence (AI), data analytics, and cybersecurity. \url{https://careertech.org/career-clusters/digital-technology/}.} represents the largest occupational cluster, comprising 28\% of adopters and 30\% of queries, followed by academia, finance, marketing, and entrepreneurship. These occupational clusters collectively account for over 70\% of total adopters and queries. They also tend to have higher agent adopter or agentic query shares than their user shares.

\subsubsection*{Use cases}
\begin{figure}[H]
    \centering
    \includegraphics[width=.8\linewidth]{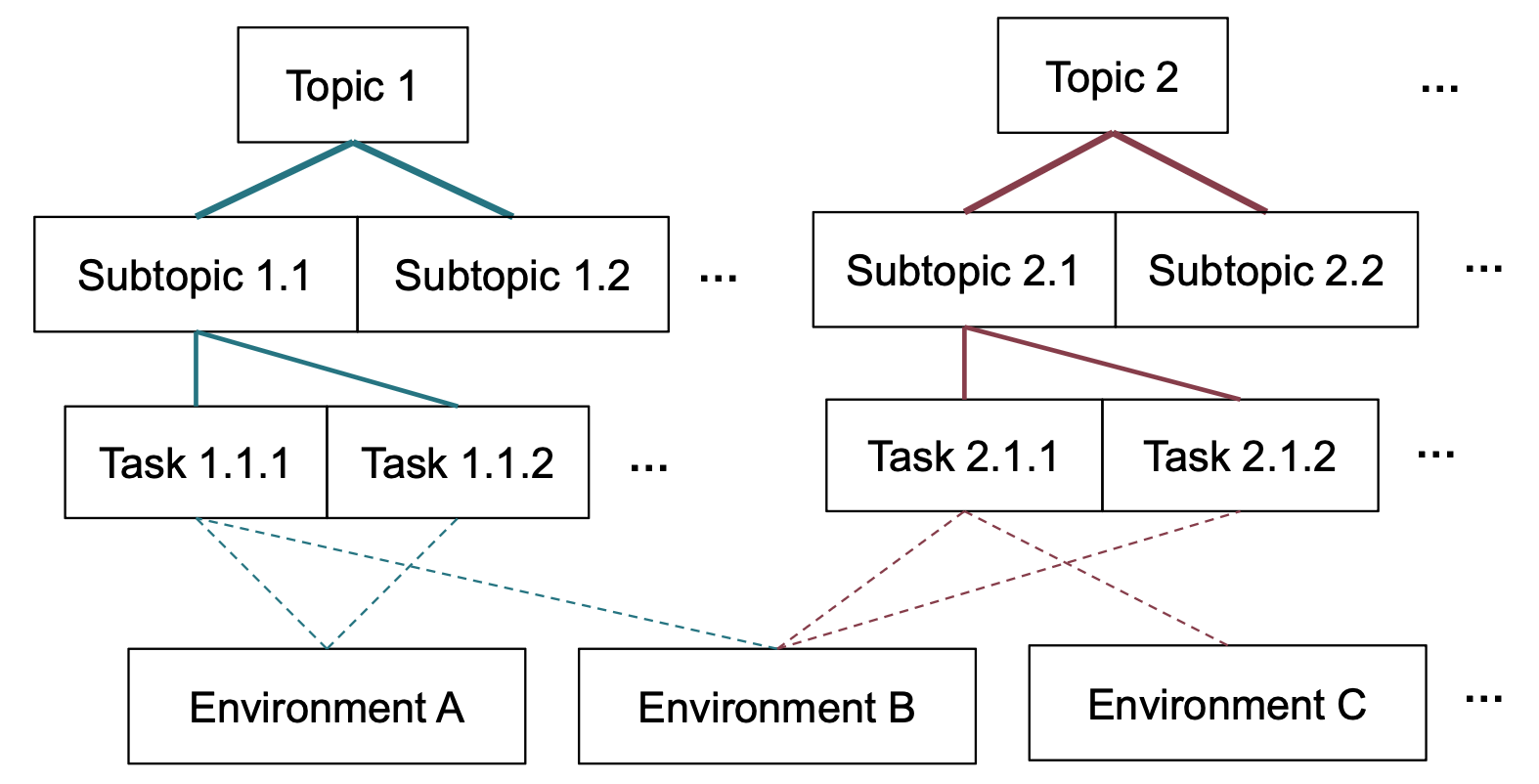}

      \begin{minipage}[t]{.8\textwidth}
        \raggedright
        \scriptsize
        Note: A topic contains multiple subtopics; a subtopic contains multiple tasks; a task can be connected to multiple environments; and an environment can be connected to one or multiple tasks within or across different topics and subtopics. The colors indicate different topics, solid lines indicate connections between topics, subtopics, and tasks, and dashed lines indicate which tasks are performed in which environments. Note that task 1.1.1 can be performed in environments A and B, whereas task 1.1.2 can only be performed in environment B. Tasks 2.1.1 and 2.1.2 indicate similar patterns. Environment B spans subtopics 1.1 and 2.1, whereas environments A and C are specific to a single subtopic. We substantiate the structure with some examples. The query ``unsubscribe me from all promotional emails that I receive more than twice per month'' would be labeled as \{Topic: Productivity \& Workflow, Subtopic: Email Management, Task: Search or filter emails, Delete or unsubscribe emails\}. In this case, searching or filtering emails, and deleting or unsubscribing from them, can both be performed in environments such as Gmail or Outlook. Now imagine another query that gets classified into \{Topic: Shopping \& Commerce, Subtopic: Shopping for Goods, Task: Search discounts, Make product purchase\}; both tasks can be performed on Instacart, whereas only search discounts can be performed on SimplyCodes, as it only shows discount codes and does not sell products directly. Facebook is one example of a cross-topic environment—it could be the environment for Media \& Entertainment queries, but also for Shopping \& Commerce queries when they are about products listed on Facebook Marketplace.
      \end{minipage}

    \caption{Hierarchical Structure of the Agentic Taxonomy}
    \label{fig:taxonomy}
\end{figure}

We illustrate the hierarchical structure of our agentic taxonomy in Figure \ref{fig:taxonomy} and report our complete taxonomy in Table \ref{tab:taxonomy}. Productivity is the dominant topic with a 36\% share. It is followed by learning (21\%), media (16\%), and shopping (10\%). The most prevalent subtopics with over 5\% query share include courses (13\%), goods shopping (9\%), research (8\%), document editing (8\%), account management (7\%), and social media (7\%). The most frequently observed tasks are exercise assistance (9\%), research information summarization and analysis (7\%), document creation and editing (7\%), product search and filtering (6\%), and research information search and filtering (6\%). We also study the use of agents across environments, which are the websites on which these tasks are performed. The concentration of environments varies substantially across subtopics: the top 5 environments account for 97\% of queries in music, 97\% in videos, and 96\% in professional networking, compared to only 28\% in account management, 35\% in shopping for services, and 37\% in project management. Across all use cases, 55\% of agentic queries originate from personal use settings, 30\% from professional use settings, and 16\% from educational use settings. In the short term, users show strong within-topic persistence, demonstrating stickiness in use cases; when topic transitions occur, they are more likely to migrate toward productivity, learning, or media topics. Over time, query shares shift from travel and media topics to productivity, learning, and career topics. 

\clearpage
\vspace*{\fill}
\vspace*{\fill}

\clearpage
\vspace*{\fill}
\begin{table}[H]
\centering
\scriptsize
\begin{threeparttable}
\begin{tabular}{|>{\centering\arraybackslash}m{3cm}|>{\centering\arraybackslash}m{3.5cm}|>{\centering\arraybackslash}m{8cm}|}
\hline
\textbf{Topics} & \textbf{Subtopics} & \textbf{Tasks} \\ 
\hline
\multirow{9}{3cm}{\centering Productivity \& Workflow} & Account Management & Register/log in to accounts, Manage settings/profiles, Manage files, Summarize/analyze account information \\ 
\cline{2-3}
& Document \& Form Editing & Create/edit documents, Search/filter documents, Summarize/analyze documents \\ 
\cline{2-3}
& Multimedia Editing & Create/edit multimedia, Search/filter multimedia, Summarize/analyze multimedia \\ 
\cline{2-3}
& Email Management  & Search/filter emails, Create/edit emails, Send emails, Delete/unsubscribe emails, Summarize/analyze emails \\ 
\cline{2-3}
& Spreadsheet \& Data Editing & Create/edit spreadsheets, Search/filter spreadsheets, Summarize/analyze spreadsheets \\ 
\cline{2-3}
& Computer Programming & Create/edit code, Execute code, Summarize/analyze code \\ 
\cline{2-3}
& Investments \& Banking & Search/filter stocks, Buy/sell stocks, Summarize/analyze investment information, Summarize/analyze banking information \\ 
\cline{2-3}
& Project Management & Create/edit projects, Summarize/analyze project information \\ 
\cline{2-3}
& Calendar Management & Create/edit events, Check availability, Search/filter events, Summarize/analyze events \\ 
\hline
\multirow{2}{3cm}{\centering Learning \& Research} & Courses & Navigate courses, Summarize/analyze course materials, Assist exercises \\ 
\cline{2-3}
& Research & Search/filter research information, Summarize/analyze research information \\ 
\hline
\multirow{6}{3cm}{\centering Media \& Entertainment} & Social Media \& Messaging & Search/filter social media posts, Summarize/analyze social media posts, Create social media posts, Engage with social media posts, Send social media/text messages \\ 
\cline{2-3}
& Online Games & Search/filter online games, Summarize/analyze online game information, Play online games \\ 
\cline{2-3}
& Movies, TV, \& Videos & Search/filter videos, Summarize/analyze videos, Play videos, Navigate within videos, Manage playlists \\ 
\cline{2-3}
& Music \& Podcasts & Search/filter music/podcasts, Summarize/analyze music/podcasts, Play music/podcasts, Manage playlists \\ 
\cline{2-3}
& News & Search/filter news, Summarize/analyze news \\ 
\cline{2-3}
& Sports & Search/filter match/player information, Summarize/analyze match/player statistics \\ 
\hline
\multirow{3}{3cm}{\centering Shopping \& Commerce} & Goods & Search/filter products, Search discounts, Summarize/analyze product information, Add products to cart, Make product purchase, Manage orders \\ 
\cline{2-3}
& Services & Search/filter products, Search discounts, Summarize/analyze product information, Add products to cart, Make product purchase, Manage orders \\ 
\hline
\multirow{4}{3cm}{\centering Travel \& Leisure} & Flights \& Transportation & Search/filter flights \& transportation, Summarize/analyze flights \& transportation, Add flights \& transportation to cart, Book flights \& transportation \\ 
\cline{2-3}
& Trip Itineraries & Search/filter destinations, Plan trips, Summarize/analyze trips \\ 
\cline{2-3}
& Lodging & Search/filter lodging, Summarize/analyze lodging information, Add lodging to cart, Book lodging \\ 
\cline{2-3}
& Restaurants & Search/filter restaurants, Summarize/analyze restaurant information, Book restaurants \\ 
\hline
\multirow{2}{3cm}{\centering Job \& Career} & Job Search \& Application & Search/filter jobs, Summarize/analyze job descriptions, Complete applications \\ 
\cline{2-3}
& Professional Networking & Search/filter professional profiles, Summarize/analyze professional profiles, Send professional connection requests/messages, Engage with professional profiles/posts \\ 
\hline
\end{tabular}
\begin{tablenotes}

\scriptsize
    \item Note: The table contains all topics, subtopics, and tasks in the agentic taxonomy, except ``Other''. Topics and subtopics are general goals, and tasks are specific tasks the agent is expected to complete to achieve those goals. A query is classified into one topic, one subtopic underneath that topic, and one or more tasks underneath that subtopic. Queries that cannot be classified into the taxonomy at a given level are labeled as ``Other'' at that level and all subsequent levels. For example, a query that does not belong to any of the topics would be labeled as ``Other'' at topic, subtopic, and task levels; a query that belongs to productivity but does not belong to any of the subtopics under productivity will be labeled as ``Other'' at subtopic and task levels; a query that belongs to productivity and email management but does not belong to any of the tasks under email management will be labeled as ``Other'' at the task level. 
\end{tablenotes}
\caption{Agentic Taxonomy—Topics, Subtopics, and Tasks} \label{tab:taxonomy}
\end{threeparttable}
\end{table}

\clearpage
The remainder of this paper is structured as follows. Section~\ref{related work} reviews related literature and highlights our contributions. Section~\ref{data} describes our sampling methodology and data privacy safeguards. Section~\ref{taxonomy} explains the development of our hierarchical agentic taxonomy. Section~\ref{result} presents our main findings on AI agent adoption patterns, usage intensity, and use cases. Finally, Section~\ref{discussion} discusses the implications of our findings for researchers, businesses, and policymakers, while acknowledging limitations and identifying promising avenues for future research that we aim to pursue. Key figures and tables are included in the main text. Additional figures, tables, and other supplementary materials are provided in the \hyperref[appendix]{Appendices}.

\section{Related Work}\label{related work}

Our paper is directly related to the literature on how people use LLMs and AI agents in real-world settings.\footnote{Following our definition of AI agents, we do not discuss papers that do not involve the agent taking actions to manipulate their environments.} Our paper extends recent work on the adoption and usage of LLM chatbots. Several prominent studies have examined this topic, including \cite{handa2025economic}, which documents user interactions with Claude, and \cite{zhao2024wildchat} and \cite{chatterji2025people}, which analyze how people use ChatGPT. Anthropic has also released detailed analyses focusing on specific user groups, such as university students \citep{handa2025education}, educators \citep{benthand2025education}, and different geographies and enterprises \citep{appelmccrorytamkin2025geoapi}. In addition, \cite{aubakirova2025stateofai} uses OpenRouter data to study LLM chatbot usage across both open- and closed-source models. These papers developed taxonomies to categorize standard Q\&A queries. We also create a taxonomy using internal data from an AI product; however, our key contribution is the focus on agentic queries. The main difference is that Q\&A queries focus on information exchange between the user and model in a conversation. In contrast, agentic queries focus on the agent executing tasks on the user's behalf in an external environment. 

Evidence on how people use AI agents in the field is limited and typically focuses on specific use cases, such as coding. For example, \cite{claudecode} studies the usage of Claude Code, a coding agent, in software development, and \cite{sarkar2025ai} investigates the adoption, usage, and productivity impact of coding agents in Cursor. Our contribution differs in that we analyze a general-purpose agent operating across all common use cases.\footnote{Although not directly related to the focus of our paper, it is worth noting adjacent research that examines the adoption and usage of LLM chatbots through user surveys (e.g., \cite{humlum2025unequal}, \cite{bick2024rapid}, \cite{handa2025interviewer}); the productivity and performance impact of LLM chatbots across various occupations and tasks through field (e.g., \cite{dell2023navigating}, \cite{wiles2024genai}, \cite{brynjolfsson2025generative}, \cite{cui2025effects}, \cite{vendraminelli2025genai}) and lab experiments (e.g., \cite{noy2023experimental}, \cite{peng2023impact}, \cite{merali2024scaling}); and the behavior of AI agents and human–agent collaboration through case studies (e.g., \cite{anthropic2025vend}), firm surveys (e.g., \cite{pan2025measuring}), or lab experiments (e.g., \cite{allouah2025your}, \cite{ju2025collaborating}).}

\section{Data}\label{data}

\subsection*{Sampling} 

Our analysis leverages three samples collected between July 9 and October 22, 2025—that is, from the launch date to 20 days after general availability. Each sample is tailored to a particular set of research questions.

We define a Comet user as a user who has made at least one query on Comet during the study period. At the user level, we exclude all enterprise users, users under the Perplexity for Government program, users who deleted their accounts during the sampling period, users who opted out of data retention for model training during that period, and logged-out users. At the query level, we define an agentic query as one that involves the agent taking control of the browser or taking actions on external applications—such as email or calendar clients—through connectors built on the Model Context Protocol (MCP) or via API calls.\footnote{\url{https://www.anthropic.com/news/model-context-protocol}} Under this stricter definition, we do not consider all queries with tool use (such as web search or code interpreter) as agentic, since these tools merely exchange information with external environments but do not manipulate them. When users onboard onto Comet, sample agentic queries are shown for demonstration purposes; we remove these queries to focus only on user-initiated ones. In rare cases, a single agentic query might trigger multiple browser-control, MCP, or API calls; we exclude such cases to focus on queries that trigger a single call, ensuring a clean inference of user intent. Lastly, we exclude queries made in Comet's incognito mode. We describe the three samples we analyze below.

\subsubsection*{Sample A: The population of Comet users and queries}\label{all sample}

We use the entire population of millions of users and hundreds of millions of queries on Comet—both agentic and non-agentic—during the sampling period to understand overall patterns in adoption and usage intensity. 

\subsubsection*{Sample B: A random sample of Comet users and queries}\label{user sample}
We draw random samples of 100,000 Comet users and their recent queries—both agentic and non-agentic—during the study period to infer their O*NET occupation clusters and subclusters, enabling us to examine variation in adoption and usage intensity across occupations.\footnote{\url{https://www.onetonline.org/find/career?c=0}} The sampling includes two stages. First, a random set of users is selected, then for each user, a random set of queries from recent dates is selected. The sampled queries are then concatenated into a single string and labeled using a classifier against the occupation taxonomy. We include university students as a separate cluster as they are not included in the occupation taxonomy.\footnote{Students are treated as a distinct cluster, separate from the education cluster, which is reserved for education professionals. Student status is verified through a third-party vendor.} 

\subsubsection*{Sample C: A random sample of Comet agent users and all their agentic queries}\label{agent sample}
We draw another random sample of 100,000 agent users and classify all their agentic queries using a novel agentic taxonomy to identify common use cases. The sampling is performed only at the user level: once a user is selected, all their agentic queries are included in the analysis. This procedure allows us to track within-user agent usage trajectories and uncover longitudinal patterns. In large user samples, the queries are also representative of query-level estimands, including common use cases. For the same sample of users, we further infer their O*NET occupation clusters and subclusters.

\subsection*{Data Privacy}

We follow industry standards and implement multiple safeguards to ensure that no human uses any personally identifiable information (PII) at any point in the analysis.\footnote{For more details on Perplexity's privacy policy, see \url{https://www.perplexity.ai/hub/legal/privacy-policy}.} 

First, as noted above, enterprise users, users under the Perplexity for Government program, users who deleted their accounts during the sampling period, users who opted out of data retention for model training during that period, logged-out users, and queries made in incognito mode are excluded from the analysis. Second, our analysis does not use any demographic information, names, email addresses, or other real-world identifiers; all user-level matching is performed through internal numerical user IDs. Third, we employ automated classifiers to label occupations and use cases. The classifier input is not the raw query text but a reformulated description of the underlying intent, enriched with context such as prior queries in the same conversation and the website on which the query was made. Lastly, all results reported in the paper are presented only in a highly aggregated form.

\section{Agentic Taxonomy}\label{taxonomy}

We develop a hierarchical agentic taxonomy guided by two principles. First, it should comprehensively capture common agentic intents so that it can generalize to other agentic products beyond Comet. Second, it should have a hierarchical structure that reveals higher-level goals while distinguishing specific lower-level tasks and actions. 

To achieve these goals, we adopt a bottom-up approach consisting of three phases: exploration, refinement, and classification. In the exploration phase, we draw a random sample of agentic queries, extract their embeddings, and apply K-means clustering to group them based on semantic similarity. Queries are then sampled from each cluster and concatenated into a single string representing that cluster. We then summarize each concatenated string to interpret the meaning of each cluster. In the refinement phase, we manually examine the cluster labels identified in the previous step to merge, split, trim, or expand them, following the guiding principles. When a significant share of queries is labeled as ``Other'', suggesting that the provided taxonomy does not sufficiently capture them, we re-classify the queries in that cluster using the bottom-up approach in the first step to identify clusters missing from the taxonomy and update it. Finally, we classify agentic queries within the finalized taxonomy using a query classification model. 

Our final taxonomy consists of three hierarchical levels—\textit{topic}, \textit{subtopic}, and \textit{task}—as illustrated in a stylized diagram in Figure \ref{fig:taxonomy}, along with their connections to the environments the tasks are performed in.\footnote{Queries
that cannot be classified into the taxonomy at a given level are labeled as “Other” at that level and all subsequent levels.
For example, a query that does not belong to any of the topics would be labeled as ``Other'' at the topic, subtopic, and task
levels; a query that belongs to productivity but does not belong to any of the subtopics under productivity will be labeled
as ``Other'' at the subtopic and task levels; a query that belongs to productivity and email management but does not belong
to any of the tasks under email management will be labeled as ``Other'' at the task level.} The full taxonomy is summarized in Table \ref{tab:taxonomy}. 

Topics and subtopics are top- and mid-level use cases of the agent, indicating the overall goal, while tasks are the specific tasks the agent is expected to complete to achieve that goal. Each query is classified into one topic, one subtopic, and one or more tasks. For instance, the query ``unsubscribe me from all promotional emails that I receive more than twice per month'' would be labeled as \{Topic: Productivity \& Workflow, Subtopic: Email Management, Task: Search or filter emails, Delete or unsubscribe emails\}. 

The environments the agent operates in are observed in the data and can be connected to our taxonomy. Tasks in a particular subtopic are performed in a specific set of environments, and each environment might involve one or more of these tasks. For instance, under the subtopic \textit{Email Management}, tasks such as \textit{Search or filter emails} and \textit{Delete or unsubscribe emails} can both be performed in environments such as Gmail or Outlook; under \textit{Shopping for Goods}, \textit{Search discounts} and \textit{Make product purchase} can both be performed on Instacart, whereas only \textit{Search discounts} can be performed on SimplyCodes as it only shows discount codes and does not sell products directly. Furthermore, an environment might cut across multiple topics and subtopics. For instance, Facebook could be the environment for \textit{Media {\rmfamily\itshape\&} Entertainment} queries, but also \textit{Shopping \& Commerce} queries when they are about products listed on Facebook Marketplace. We further categorize the usage context into \textit{personal}, \textit{professional}, and \textit{educational} domains.

We validate the classification accuracy against a golden dataset of 1,000 anonymized and desensitized queries. The classifier-assigned labels agree with the topics, subtopics, tasks, and usage context in the golden dataset 89\%, 83\%, 81\%, and 83\% of the time, respectively. More details about the validation are provided in Appendix \ref{validation}.

\section{Main Results} \label{result}

We first discuss the results on the adoption (extensive margin) and usage intensity (intensive margin), and then the use cases (agentic taxonomy). 

\subsection{Adoption and Usage Intensity}\label{adoption and usage}

We define agent adopters as users who had at least one agentic query in the sampling period. The results below are all based on Sample \hyperref[all sample]{A}, except for occupation, which is based on Sample \hyperref[all sample]{B}. 

Figure \ref{fig:adoption-usage} in Appendix \ref{figures and tables} shows that agent adoption and overall usage as measured by agentic query volumes are growing steadily over the period studied, with an increased pace after Comet became generally available. About 60\% of agent users were acquired, and 50\% of agentic queries occurred post-GA. The query volumes grow at a slightly higher rate than adopters. 

We analyze the adoption and usage patterns of user segments defined by cohort, country, and occupation. To capture the magnitude of adoption and usage within a user segment relative to its user share, we define the Perplexity Agent Adoption Ratio (AAR) and the Agent Usage Ratio (AUR) as the ratio of a segment’s agent adopter share or agentic query share to its user share.\footnote{These ratio-based metrics are often used to quantify the relative degrees of adoption and usage (e.g., \cite{appelmccrorytamkin2025geoapi})} An AAR or AUR greater than one indicates that a segment is over-represented in the adopters or queries relative to their population base, and vice versa. 

\subsubsection*{By cohort}
Table \ref{tab:adoption-usage-cohort} shows that among the three cohorts by access dates, earlier adopters (those with access before GA) account for about 30\% of total users, but about 50\% of agent adopters and 70\% of agentic queries. The disparity is more pronounced in usage intensity than in adoption—an average user in the first cohort (July 9) is twice as likely to adopt the agent but makes nine times as many agentic queries as an average user in the GA cohort (October 2). AAR and AUR both decrease in the order of access cohorts. 

In general, the composition of early adopters is endogenous to the rollout plan; in our case, it is based on the subscription tiers. Nonetheless, these results are consistent with general patterns in the adoption and usage of new technologies \citep{Moore1991Crossing}: early adopters disproportionately drive initial adoption and usage, and subsequent diffusion may require additional educational efforts. With improved agent-to-agent collaboration capabilities and standardized protocols, one might expect stronger network effects that could accelerate adoption and usage in the future.\footnote{For instance, Agent2Agent Protocol (A2A). \url{https://a2a-protocol.org/latest/}.}

\begin{table}[H]
\centering
\begin{threeparttable}
\caption{Agent Adoption and Agentic Query by Cohort}
\label{tab:adoption-usage-cohort}
\small 
\begin{tabular}{lcccccc}
\toprule
\textbf{Cohort} & \textbf{User Share (\%)} & \textbf{Agent Adopter Share (\%)}  & \textbf{Agentic Query Share (\%)} & \textbf{AAR} & \textbf{AUR} \\
\midrule
July 9 & 4.3 & 7.7  & 18.9 & 1.79 & 4.40 \\
August 13 & 28.3 & 38.1 & 48.5 &  1.35 & 1.71 \\
October 2 & 67.4 & 54.2 & 32.7 & 0.80 & 0.49 \\
\bottomrule
\end{tabular}

\begin{tablenotes}
\scriptsize
\item Note: The table shows the agent adoption and agentic query by cohort. User share is the number of users in each cohort divided by the total users. Agent adopter and query share are the numbers of adopters and agentic queries in each cohort, divided by the total number of adopters and agentic queries. AAR (Agent Adoption Ratio) is the ratio between agent adopter share and user share. AUR (Agent Usage Ratio) is the ratio between agentic query share and user share. AAR and AUR greater (less) than 1 indicate that a cohort is over-represented (under-represented) in agent adopters and queries relative to their user base.
\end{tablenotes}
\end{threeparttable}
\end{table}

\subsubsection*{By country}

\begin{figure}[H]
  \centering
  \small
  \begin{subfigure}[b]{0.7\textwidth}
    \includegraphics[width=\linewidth]{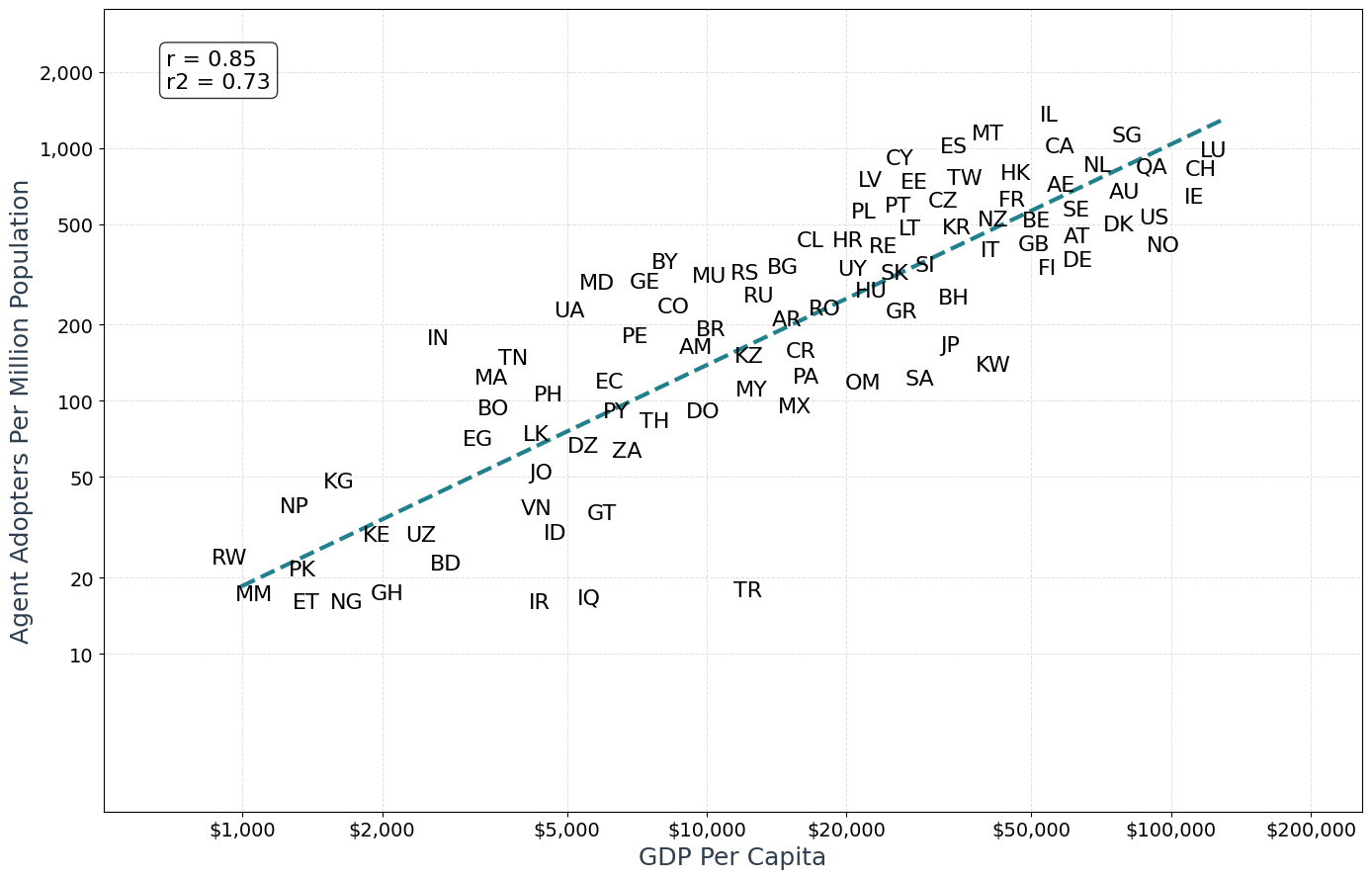}
    \caption{Log GDP Per Capita vs. Log Agent Adopters Per Million Population}
  \end{subfigure}
  
  \vspace{0.5cm} 

  \begin{subfigure}[b]{0.7\textwidth}
    \includegraphics[width=\linewidth]{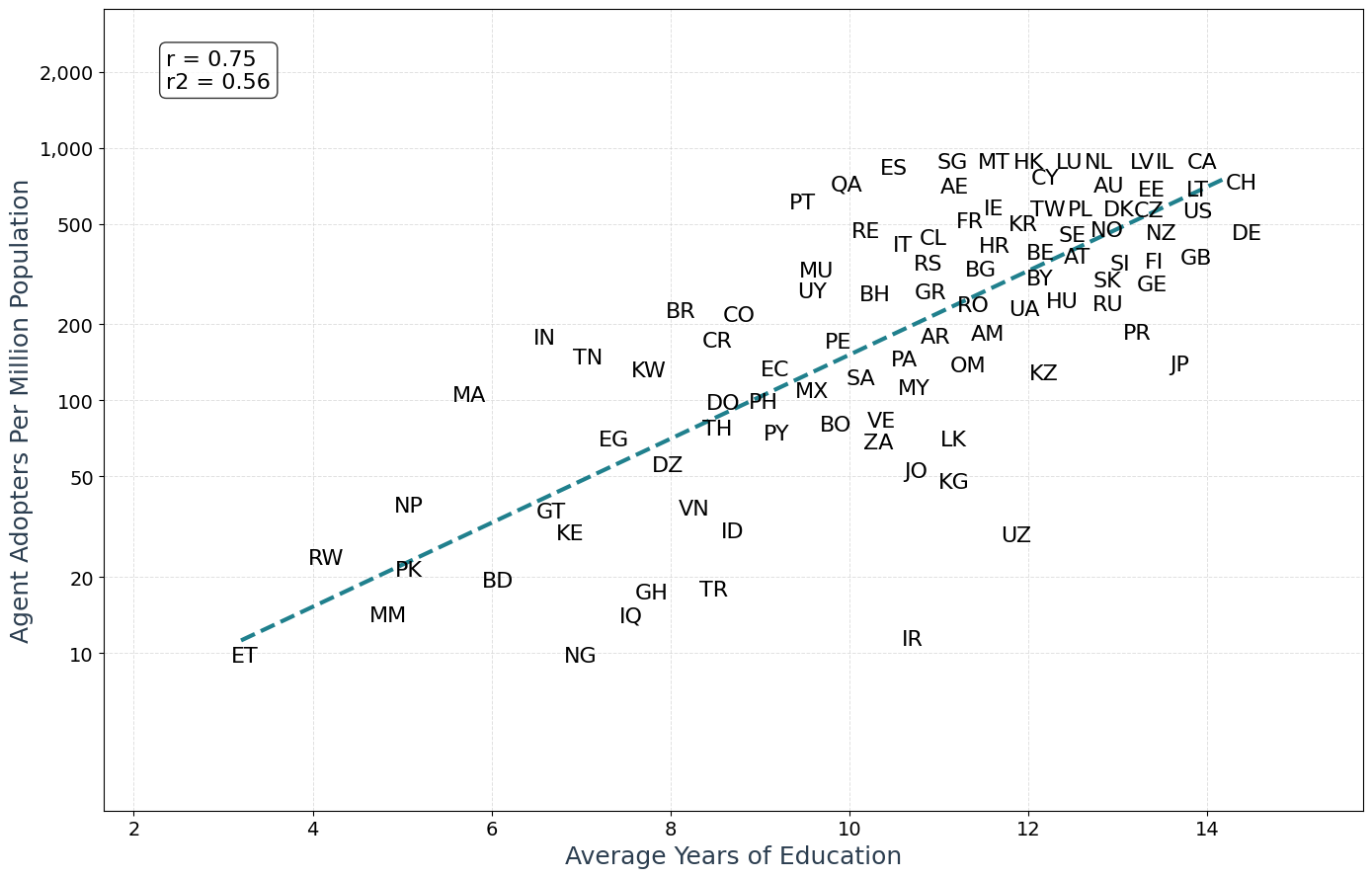}    
    \caption{Average Years of Education vs. \\Log Agent Adopters Per Million Population}
  \end{subfigure}
  \caption{Log GDP Per Capita and Average Years of Education vs. \\
  Log Agent Adopters Per Million Population by Country}
    \begin{minipage}[t]{\textwidth}
      \raggedright
      \scriptsize
      Note: The plots show the scatterplots and best-fitting lines of log GDP per capita and average years of education vs. the log of agent adopter per million population for the top 100 countries by agent adopter count. The plots are on a log scale, but the labels are in absolute values for better readability. Jitter is applied to the country labels to provide better visual separation. $r$ is the correlation coefficient, $p$ is the p-value of the regression coefficient, and $R^2$ is the R-squared of the regression lines. The GDP and population data are from World Bank World Development Indicators (2024)\footnote{\url{https://data.worldbank.org/}} and the average years of education data are from UNDP Human Development Report (2024)\footnote{\url{https://hdr.undp.org/content/human-development-report-2023-24}}.
    \end{minipage}
  \label{fig:adoption-country-gdp and edu}
\end{figure}

\begin{figure}[H]
  \centering
  \begin{subfigure}[b]{0.7\textwidth}
    \includegraphics[width=\linewidth]{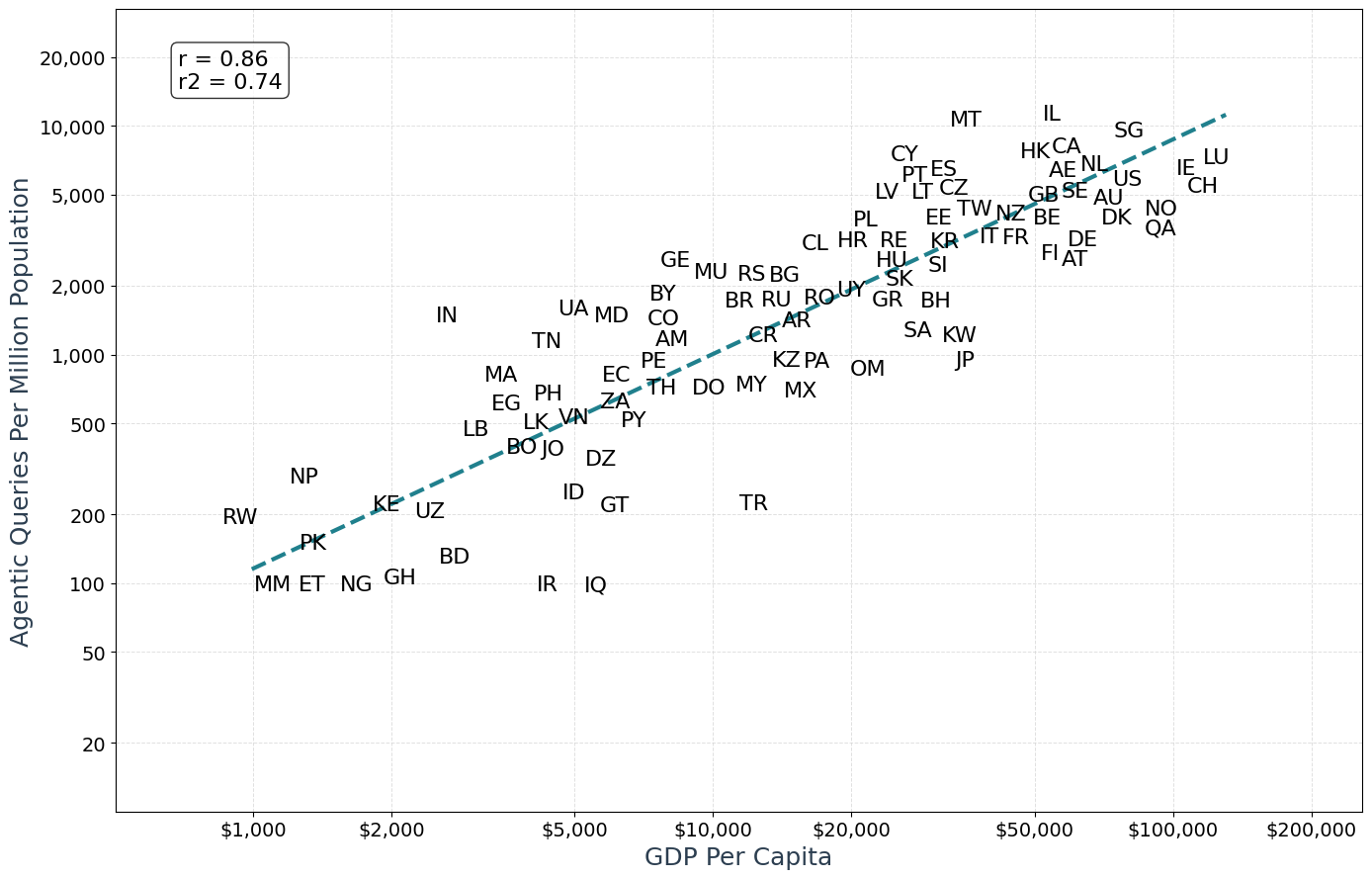}
    \caption{Log GDP Per Capita vs. Log Agentic Queries Per Million Population 
    }
  \end{subfigure}
  
  \vspace{0.5cm} 

  \begin{subfigure}[b]{0.7\textwidth}
    \includegraphics[width=\linewidth]{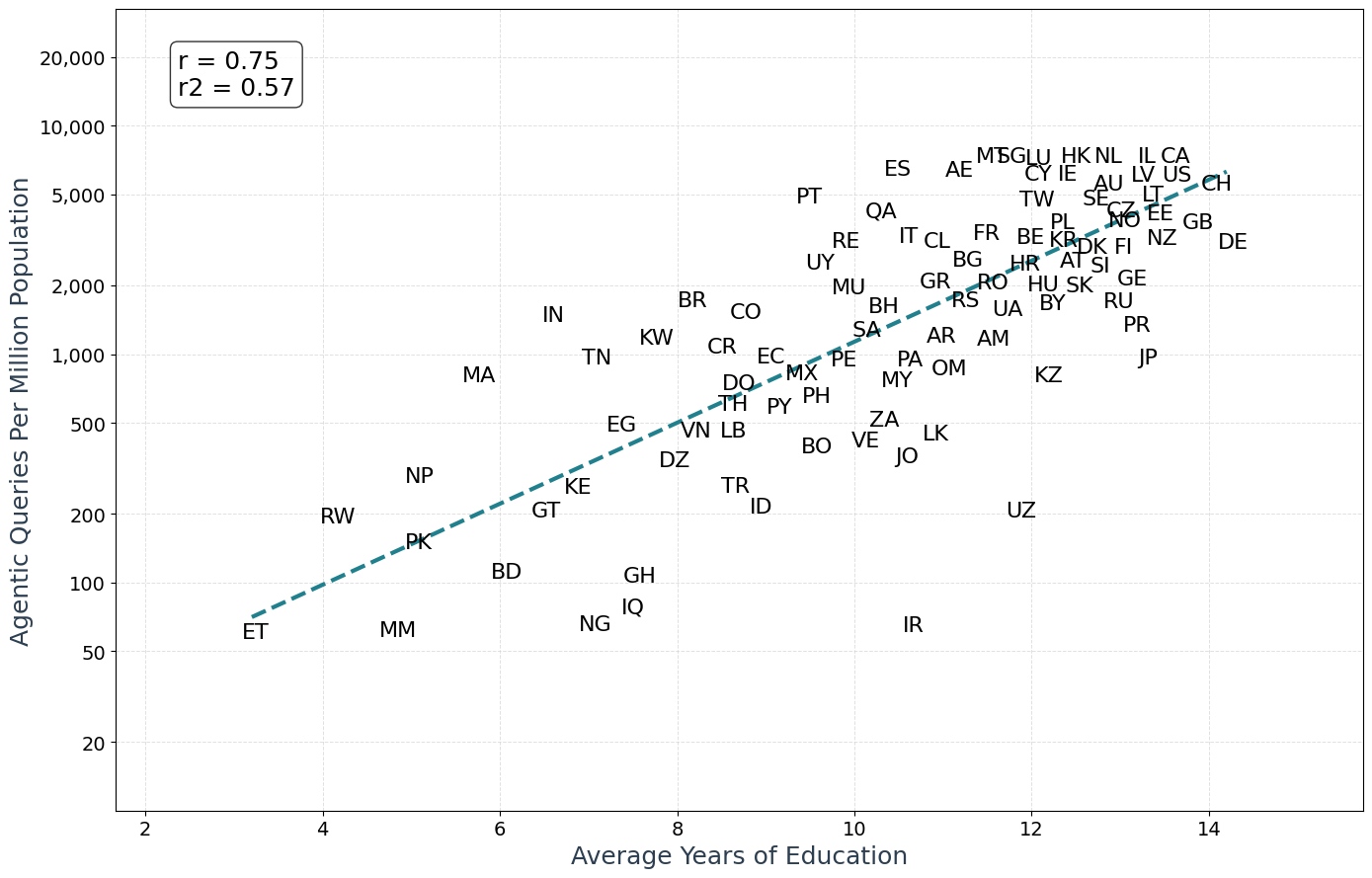}    
    \caption{Average Years of Education vs. \\ Log Agentic Queries Per Million Population
    }
  \end{subfigure}
  \caption{Log GDP Per Capita and Average Years of Education vs.\\ Log Agentic Queries Per Million Population by Country}
    \begin{minipage}[t]{\textwidth}
      \raggedright
       \scriptsize
      Note: The plots show the scatterplots and best-fitting lines of log GDP per capita and average years of education vs. the log of agentic query per million population for the top 100 countries by agentic query count. The plots are on a log scale, but the labels are in absolute values for better readability. Jitter is applied to the country labels to provide better visual separation. $r$ is the correlation coefficient, $p$ is the p-value of the regression coefficient, and $R^2$ is the R-squared of the regression lines. The GDP and population data are from World Bank World Development Indicators (2024)\footnote{\url{https://data.worldbank.org/}}, and the average years of education data are from UNDP Human Development Report (2024)\footnote{\url{https://hdr.undp.org/content/human-development-report-2023-24}}
    \end{minipage}
  \label{fig:usage-country-gdp and edu}
\end{figure}

Figure \ref{fig:adoption-country-gdp and edu} and \ref{fig:usage-country-gdp and edu} show that there are strong positive correlations between log agent adopters per million population and log GDP per capita ($r = 0.85, p < 0.001, R^2 = 0.73$) and years of education ($r = 0.75, p < 0.001, R^2 = 0.56$), where $r$ is the correlation coefficient, $p$ is the p-value of the correlation coefficient, and $R^2$ is the R-squared of the regression lines. The correlations between log agentic queries per million population and log GDP per capita ($r = 0.86, p < 0.001, R^2 = 0.74$) and years of education ($r = 0.75, p < 0.001, R^2 = 0.57$) follow a similar pattern. Together, they suggest that relatively more economically developed and educated countries tend to adopt and use the agent more.\footnote{Results remain consistent when population is replaced with working population.}

\clearpage

\subsubsection*{By occupation}

Table \ref{tab:adoption-occupation-cluster} ranks occupation clusters (including the student cluster) by user share, adopter share, and AAR. Digital technology is by far the largest cluster, accounting for 28\% of adopters, slightly higher than its user share. Academics (including the student and education clusters) and financial workers have an adopter share of more than 10\%. Workers in marketing, design, and entrepreneurship have an adopter share of more than 5\%. Clusters with lower adopter shares are typically those that require interacting with the physical environment. The hospitality cluster has the highest AAR at 1.36, although the sample size is significantly smaller; it is followed by marketing and entrepreneurship at 1.24 and 1.17, respectively. 

Table \ref{tab:usage-occupation-cluster} ranks occupation clusters by user share, query share, and AUR. The patterns track adoption closely—top clusters remain the same with slight changes in the rank. Comparing the AUR and AAR for the same cluster reveals patterns in the degree of usage intensity relative to the degree of adoption: students and workers in entrepreneurship, marketing, and digital technology all have AUR / AAR greater than one, suggesting that their tendency to use the agent conditional on adoption is even stronger than their tendency to adopt.

Tables \ref{tab:adoption-occupation-subcluster} and \ref{tab:usage-occupation-subcluster} in Appendix \ref{figures and tables} show the top 10 occupation subclusters and their user shares by agent adopter shares and AAR, and agentic query share and AUR, respectively. Patterns at the subcluster level are largely consistent with those at the cluster level, with software engineers being the largest subcluster, accounting for 14\% of adopters and 15\% of queries, and having AAR and AUR around 1.1 and 1.2, respectively. All other subclusters are below 6\% in both adopter and query shares. Subclusters in marketing—such as business development and sales, digital marketing, and market research—and in entrepreneurship—such as information management, operations, and strategy—tend to have the highest AAR and AUR.

These results may reflect differences in the task composition of each occupation and how closely those tasks align with common agent use cases, which the next section examines.

\clearpage
\vspace*{\fill}

\begin{table}[H]
\centering
\begin{threeparttable}[h]
\scriptsize
\begin{tabular}{lcccccc}
\toprule

\multicolumn{6}{c}{Agent Adoption by Occupation Cluster: By AAR} \\
\midrule
\textbf{Cluster} & \textbf{User Share (\%)} & \textbf{User Share Rank} & \textbf{Agent Adopter Share (\%)} & \textbf{Agent Adopter Share Rank} & \textbf{AAR} \\
\midrule
Hospitality, Events, \& Tourism & 2.5 & 11 & 3.4 & 9 & 1.36 \\
Marketing \& Sales              & 7.2 & 6 & 8.9 & 4 & 1.24 \\
Management \& Entrepreneurship  & 6.5 & 7 & 7.6 & 6 & 1.17 \\
Digital Technology              & 26.4 & 1 & 27.7 & 1 & 1.05 \\
Supply Chain \& Transportation  & 2.2  & 13 & 2.3 & 12  & 1.05 \\
Financial Services              & 10.0 & 3 & 10.1 & 3 & 1.01 \\
Student                         & 12.4 & 2 & 12.4 & 2 & 1.00 \\
Construction                    & 2.3  & 12 & 2.2 & 13 & 0.96 \\
Energy \& Natural Resources     & 0.9  & 14 & 0.8 & 14 & 0.89 \\
Arts, Entertainment, \& Design  & 9.1  & 4 & 8.0  & 5 & 0.88 \\
Education                       & 7.7  & 5 & 6.4  & 7 & 0.83 \\
Healthcare \& Human Services    & 4.9  & 8 & 4.0  & 8 & 0.82 \\
Advanced Manufacturing          & 3.5  & 9 & 2.8  & 10 & 0.80 \\
Public Service \& Safety        & 3.4  & 10 & 2.7  & 11 & 0.79 \\
Agriculture                     & 0.9  & 15 & 0.7  & 15 & 0.78 \\
\bottomrule
\end{tabular}
\begin{tablenotes}
 \scriptsize
\item Note: The table shows agent adoption by O*NET occupation cluster. The ``Other'' category is removed. We put students in a separate cluster and educators in the Education cluster. User share is the number of users in each cluster divided by the total users. Agent adopter share is the number of adopters in each cluster divided by the total adopters. AAR (Agent Adoption Ratio) is the ratio between agent adopter share and user share. AAR greater (less) than 1 indicates that a cluster is over-represented (under-represented) in agent adoption relative to their user base.
\end{tablenotes}
\caption{Agent Adoption by Occupation Cluster}
\label{tab:adoption-occupation-cluster}
\end{threeparttable}
\end{table}

\begin{table}[H]
\centering
\begin{threeparttable}
\centering
\footnotesize
\begin{tabular}{lccccc}
\toprule

\multicolumn{6}{c}{Agentic Query by Occupation Cluster: By AUR} \\
\midrule
\textbf{Cluster} & \textbf{User Share (\%)} & \textbf{User Share Rank} & \textbf{Agentic Query Share (\%)} & \textbf{Agentic Query Share Rank} & \textbf{AUR} \\
\midrule
Marketing \& Sales              & 7.2  & 6 & 10.5 & 3 & 1.46 \\
Management \& Entrepreneurship  & 6.5  & 7 & 9.0  & 4 & 1.38 \\
Student                         & 12.4 & 2 & 15.6 & 2 & 1.26 \\
Digital Technology              & 26.4 & 1 & 29.6 & 1 & 1.12 \\
Hospitality, Events, \& Tourism & 2.5  & 11 & 2.6  & 9 & 1.04 \\
Supply Chain \& Transportation  & 2.2  & 13 & 2.0  & 10 & 0.91 \\
Financial Services              & 10.0 & 3 & 8.6  & 5 & 0.86 \\
Arts, Entertainment, \& Design  & 9.1  & 4 & 6.9  & 6 & 0.76 \\
Education                       & 7.7  & 5 & 5.6  & 7 & 0.73 \\
Construction                    & 2.3  & 12 & 1.6  & 13 & 0.70 \\
Healthcare \& Human Services    & 4.9  & 8 & 3.3  & 8 & 0.67 \\
Agriculture                     & 0.9  & 15 & 0.5  & 15 & 0.56 \\
Energy \& Natural Resources     & 0.9  & 14 & 0.5  & 14 & 0.56 \\
Advanced Manufacturing          & 3.5  & 9 & 1.9  & 11 & 0.54 \\
Public Service \& Safety        & 3.4  & 10 & 1.8  & 12 & 0.53 \\
\bottomrule
\end{tabular}

\begin{tablenotes}
 \scriptsize
\item Note: The table shows usage intensity by O*NET occupation cluster. The ``Other'' category is removed. We put students in a separate cluster and educators in the Education cluster. User share is the number of users in each cluster divided by the total users. Agent query share is the number of agentic queries in each cluster divided by the total agentic queries. AUR (Agent Usage Ratio) is the ratio between agentic query share and user share. AUR greater (less) than 1 indicates that a cluster is over-represented (under-represented) in agent usage relative to their user base.
\end{tablenotes}
\caption{Agentic Query by Occupation Cluster}
\label{tab:usage-occupation-cluster}
\end{threeparttable}
\end{table}

\vspace*{\fill}
\clearpage

\subsection{Use Cases}\label{use cases}
We document the use cases by topic, subtopic, task, environment (the websites on which the tasks are performed), and usage context. All results are based on Sample \hyperref[all sample]{C}. 

\subsubsection*{Topics and subtopics}
\begin{figure}[H]
    \centering

    \includegraphics[width=\linewidth]{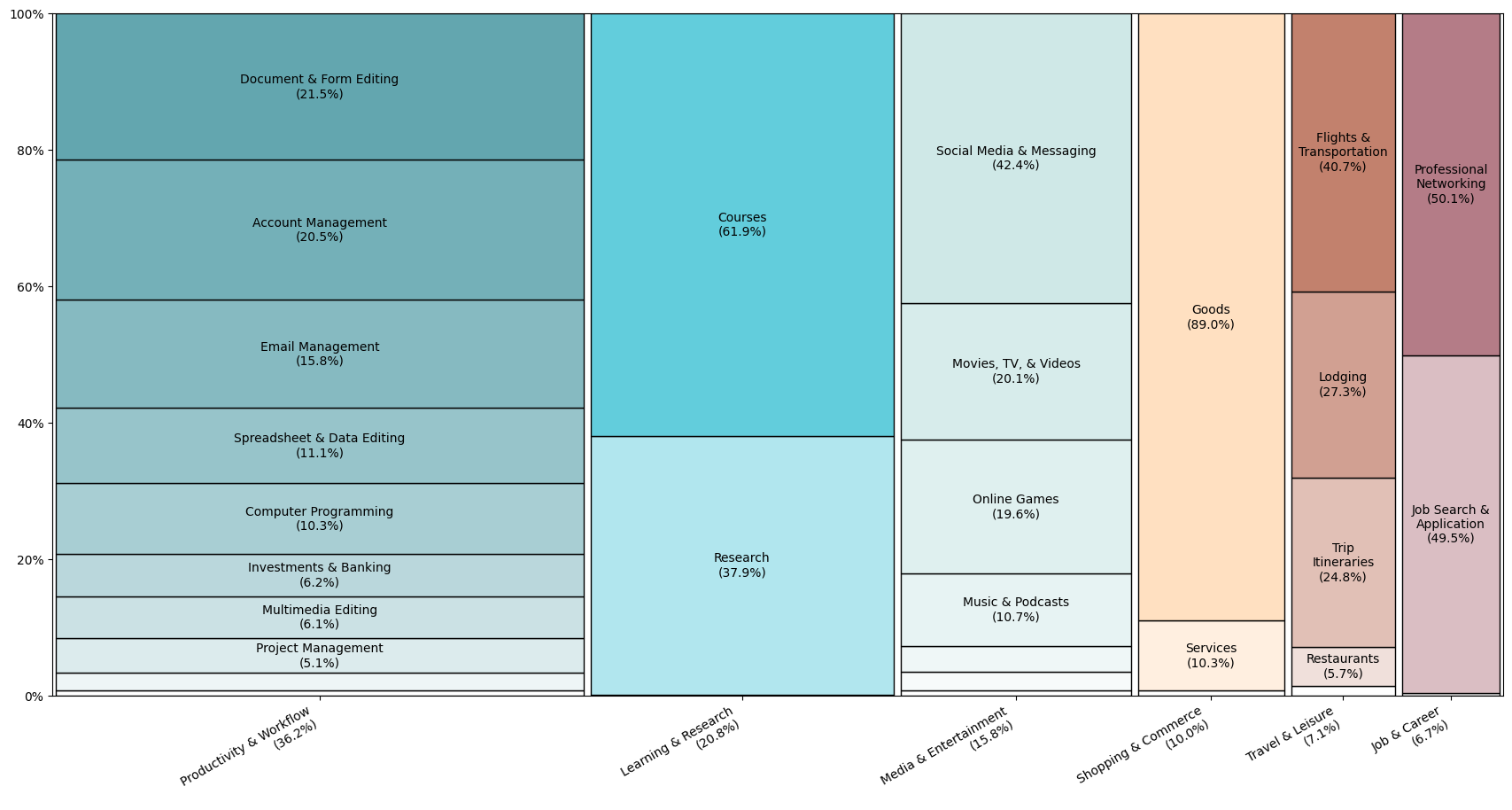}

    \caption{Topic Breakdown by Subtopic Percentage}

    \begin{minipage}[t]{\textwidth}
      \raggedright
      \footnotesize
      Note: The plot shows the percentage shares of subtopics within each topic. Bar width is proportional to topic percentage, and box height within each bar is proportional to subtopic percentage. ``Other'' category (3.4\%) is not shown among the topics. The labels for subtopics that account for less than 5\% within a topic are suppressed. Topic shares are shown in the labels on the x-axis. The subtopics within each topic are sorted by percentage in decreasing order from top to bottom. The darker shades within a topic represent subtopics with higher percentages.
    \end{minipage}
    \label{fig:use case-topic-subtopic}
\end{figure}

Topics and subtopics capture the high-level goals of an agentic query. Figure \ref{fig:use case-topic-subtopic} presents the topic share and subtopic share breakdown by topic. Productivity is the largest category, accounting for 36\% of all agentic queries. Learning, media, and shopping are the other topics with over 10\% query share. The two largest topics—productivity and learning—together account for 57\% of all queries. Some topics, such as learning, shopping, and career, have a dominant subtopic that accounts for more than half of the queries in that topic. Table \ref{tab:use case-topic-subtopic} in Appendix \ref{figures and tables} also shows the overall query share of each subtopic. Courses account for 13\% of all queries, followed by goods shopping, document editing, account management, social media, and email.

\clearpage
Table \ref{tab:use case-occupation} shows the topic distribution by occupation cluster. Topic-wise, productivity remains the largest topic for most occupations, whereas learning and research is the largest for students and educators, and travel is the largest for the hospitality cluster. Occupation-wise, workers in finance have the largest query share in productivity, students have the largest share in learning, designers have the largest share in media, workers in advanced manufacturing have the largest share in shopping, workers in entrepreneurship have the largest share in career, and workers in hospitality have the largest share in travel.

\begin{table}[H]
\centering

\begin{threeparttable}[H]
\tiny
\begin{tabular}{lcccccc}
\hline
\textbf{Cluster / Topic} & \textbf{Productivity \&} & \textbf{Learning \&}  & \textbf{Media \&}  & \textbf{Shopping \&}  & \textbf{Job \&}  & \textbf{Travel \&}  \\
& \textbf{Workflow (\%)} & \textbf{Research (\%)} & \textbf{Entertainment (\%)} & \textbf{Commerce (\%)} & \textbf{Career (\%)} & \textbf{Leisure (\%)} \\
\hline
Digital Technology & 41.0 & 19.9 & 14.7 & 9.3 & 9.1 & 6.0 \\
Student & 29.1 & 43.3 & 10.6 & 5.3 & 8.1 & 3.7 \\
Management \& Entrepreneurship & 45.8 & 13.7 & 12.3 & 9.8 & 12.2 & 6.2 \\
Marketing \& Sales & 37.5 & 12.1 & 23.8 & 14.0 & 8.2 & 4.5 \\
Financial Services & 46.7 & 15.2 & 14.0 & 9.8 & 5.7 & 8.6 \\
Education & 34.1 & 37.0 & 13.8 & 6.6 & 2.6 & 5.8 \\
Arts, Entertainment, \& Design & 39.4 & 12.1 & 25.0 & 11.3 & 6.0 & 6.2 \\
Healthcare \& Human Services & 38.6 & 23.3 & 14.3 & 10.2 & 5.9 & 7.7 \\
Advanced Manufacturing & 30.4 & 19.8 & 11.8 & 25.4 & 5.4 & 7.1 \\
Public Service \& Safety & 39.7 & 26.2 & 17.7 & 7.2 & 3.2 & 6.0 \\
Hospitality, Events, \& Tourism & 29.9 & 6.6 & 13.4 & 12.2 & 2.7 & 35.2 \\
Supply Chain \& Transportation & 40.3 & 13.4 & 12.5 & 18.6 & 5.3 & 10.0 \\
Construction & 39.1 & 14.9 & 14.1 & 16.3 & 7.7 & 7.9 \\
Energy \& Natural Resources & 42.9 & 18.8 & 13.1 & 10.4 & 4.9 & 9.8 \\
Agriculture & 41.5 & 20.5 & 13.9 & 12.4 & 5.4 & 6.3 \\
\hline
\end{tabular}
\begin{tablenotes}
\scriptsize
\item Note: The table shows the distribution of topics by occupation cluster. Topic percentage ($P(\text{Topic | Occupation})$) is the topic share among all agentic queries from a given occupation cluster. Percentages may not sum to 100\% due to rounding.
\end{tablenotes}
\caption{Topic Distribution by Occupation Cluster}
\label{tab:use case-occupation}
\end{threeparttable}
\end{table}

We also examine transition patterns between consecutive queries. Figure \ref{fig:use case-topic transition} in Appendix \ref{figures and tables} shows the transition probability matrix from the previous query to the next query for all query pairs, aggregated at the user level. The results show that in most cases, topics transition into themselves, suggesting the stickiness of agent use cases. Productivity, learning, and career topics are the most sticky, whereas travel is the least sticky; media and shopping topics fall in between. When cross-topic transitions occur, they most likely transition into productivity, learning, or media topics. Then we compare users' first queries—their entry points into the agent—and all queries. Figure \ref{fig:use case-first vs all} in Appendix \ref{figures and tables} contrasts the topic distribution among the first agentic query for each user with the overall distribution. Over time, the share of queries on productivity, learning, and career topics has increased, suggesting a shift toward more cognitively oriented use cases. 

\subsubsection*{Tasks}
Tasks under topics and subtopics capture the low-level tasks the agent is expected to complete to achieve the end goals. We show the top 10 tasks in Table \ref{tab:use case-top task}. Half of the top 10 are in learning, including various learning and research assistance. The other five are split across productivity (edit documents and manage account settings), shopping (search or summarize product information), and media (search social media). 

\begin{table}[H]
\centering
\begin{threeparttable}
\footnotesize
\caption{The Top 10 Tasks}
\label{tab:use case-top task}
\begin{tabular}{l l l c}
\toprule
\textbf{Topic} & \textbf{Subtopic} & \textbf{Task} & \textbf{Overall (\%)} \\
\midrule
Learning \& Research & Courses & Assist exercises & 9.41 \\
Learning \& Research & Research & Summarize/analyze research information & 6.71 \\
Productivity \& Workflow & Document \& Form Editing & Create/edit documents/forms & 6.58 \\
Shopping \& Commerce & Goods & Search/filter products & 6.43 \\
Learning \& Research & Research & Search/filter research information & 5.95 \\
Shopping \& Commerce & Goods & Summarize/analyze product information & 5.18 \\
Productivity \& Workflow & Account Management & Manage settings/profiles & 4.33 \\
Learning \& Research & Courses & Summarize/analyze course materials & 3.69 \\
Learning \& Research & Courses & Navigate courses & 3.31 \\
Media \& Entertainment & Social Media \& Messaging & Search/filter social media posts/messages & 3.29 \\
\bottomrule
\end{tabular}
\begin{tablenotes}
\footnotesize
\item Note: The table shows the top 10 tasks among all agentic queries. $P(\text{Task Overall}) = P(\text{Topic, Subtopic, Task}) = P(\text{Topic}) \times P(\text{Subtopic | Topic}) \times P(\text{Task | Topic, Subtopic}).$
\end{tablenotes}
\end{threeparttable}
\end{table}

Tables \ref{tab:use case-productivity-task}, \ref{tab:use case-learning-task}, \ref{tab:use case-media-task}, \ref{tab:use case-shopping-task}, \ref{tab:use case-job-task}, and \ref{tab:use case-travel-task} in Appendix \ref{figures and tables} show the main tasks under each topic and subtopic with over 5\% query shares within a subtopic. Note that because query share measures the fraction of queries in which a task is present and a query might contain multiple tasks, the task percentages under each subtopic can add up to over 100. A few subtopics contain a dominant task that appears in over 80\% of all queries in that subtopic. For instance, searching for flights and lodging both appear in 93\% of queries in the flight and lodging subtopics; editing documents and summarizing research information both appear in 85\% of queries in the document editing and research subtopics, respectively. In contrast, some subtopics show more dispersed task distributions: for instance, searching videos—the top task in the video subtopic—appears in only 48\% of all queries in that category; searching email—the top task in the email subtopic—appears in only 49\% of all queries in that category. 

Table \ref{tab:use case-task-occupation} in Appendix \ref{figures and tables} shows the top 5 tasks in each occupation cluster. In general, research, document editing, and shopping-related tasks appear consistently across clusters. Some occupation clusters feature a prominent task. For instance, search products appear in 21\% of queries from the advanced manufacturing cluster. Other clusters, in contrast, have a more diffuse task composition. The top tasks in the entrepreneurship and design clusters—summarizing research information and searching products—appear in fewer than 8\% of their queries. The top tasks by occupation also shed light on why certain occupations tend to adopt and use the agent more. Knowledge-intensive sectors such as digital technology, entrepreneurship, finance, and academia tend to use the agent for research and learning-related tasks. In contrast, highly digitized sectors such as marketing and design tend to use the agent for media-related tasks. 

\subsubsection*{Environments}
Environment refers to the external world with which the agent interacts while performing a task to achieve its goals. In our context, the environment is the specific website on which the agent operates for a given query. We show the top environments by overall query shares in Table \ref{tab:use case-top env}. The top 16 environments together account for 64\% of queries, with the top 5 alone representing 43\%. These environments are typically the dominant websites within their respective domains.

We break down the top 5 environments under each subtopic and their query shares in Tables \ref{tab:use case-productivity-env}, \ref{tab:use case-learning-env}, \ref{tab:use case-media-env}, \ref{tab:use case-shopping-env}, \ref{tab:use case-job-env}, and \ref{tab:use case-travel-env} in Appendix \ref{figures and tables}. A single environment dominates some subtopics. For instance, linkedin.com alone accounts for 93\% of queries in professional networking, and the query shares of youtube.com and docs.google.com\footnote{docs.google.com includes Google Docs, Sheets, Slides, and Forms.}—the top environments in video and spreadsheet editing subtopics—are more than twenty times larger than the share of the second environments in those subtopics. On the other hand, there is only a 2\% difference between coursera.org and netacad.com under courses, and a 3\% difference between instagram.com and x.com under social media. Table \ref{tab:use case-env-concentration} in Appendix \ref{figures and tables} shows the combined shares of the top 5 environments in each subtopic. A higher share indicates the agent usage is more concentrated in a few environments. The level of concentration varies significantly: the top 5 environments account for 97\% of queries in music, 97\% in videos, and 96\% in professional networking, compared to only 28\% in account management, 35\% in services shopping, and 37\% in project management, respectively. Lastly, Table \ref{tab:use case-env-occupation} in Appendix \ref{figures and tables} shows the top 5 environments for each occupation cluster, which are closely related to the main use cases for each occupation. 

\begin{table}[H]
\centering
\begin{threeparttable}
\small
\begin{tabular}{lr}
\toprule
\textbf{Environment} & \textbf{Overall (\%)} \\
\midrule
docs.google.com & 11.97 \\
email services combined & 11.23 \\
linkedin.com & 9.42 \\
youtube.com & 7.03 \\
amazon.com & 3.46 \\
instagram.com & 2.56 \\
messenger services combined & 2.47 \\
maps.google.com & 2.20 \\
coursera.org  & 2.04 \\
x.com  & 2.00 \\
github.com & 1.85 \\
facebook.com & 1.77 \\
netacad.com & 1.75 \\
canva.com & 1.49 \\
canvas.com & 1.44 \\
notion.so & 1.13 \\
\bottomrule
\end{tabular}
\caption{Top Agent Environments}
\label{tab:use case-top env}
\begin{tablenotes}
\scriptsize
\item Note: The table shows all environments with a query share above 1\% among all agentic queries. docs.google.com includes Google Docs, Sheets, Slides, and Forms. All email domains are grouped into ``email services combined'' and all online messengers are grouped into ``messenger services combined''. 
\end{tablenotes}
\end{threeparttable}
\end{table}

\subsubsection*{Usage context}
Lastly, we investigate agent usage across personal, professional, and educational contexts.\footnote{Note that our use case taxonomy is orthogonal mainly to the usage context. For instance, a user may ask the agent to reply to an email from a friend (personal), a colleague (professional), or a professor (educational). Similarly, users may ask the agent to shop for personal items, workplace equipment, or school supplies.} Personal use comprises about 55\% of total agentic queries, with professional and educational contexts representing 30\% and 16\%, respectively. There is a slight increase in the share of educational use and a slight decrease in the share of personal use over time, while the share of professional use remains stable. This could be driven by the public launch time of Comet overlapping with the start of the fall semester and by on-campus promotional efforts, such as early access for university students, rather than by a systematic shift in the underlying composition of user groups and use cases. 

\begin{figure}[H]
    \centering
    \includegraphics[width=\linewidth]{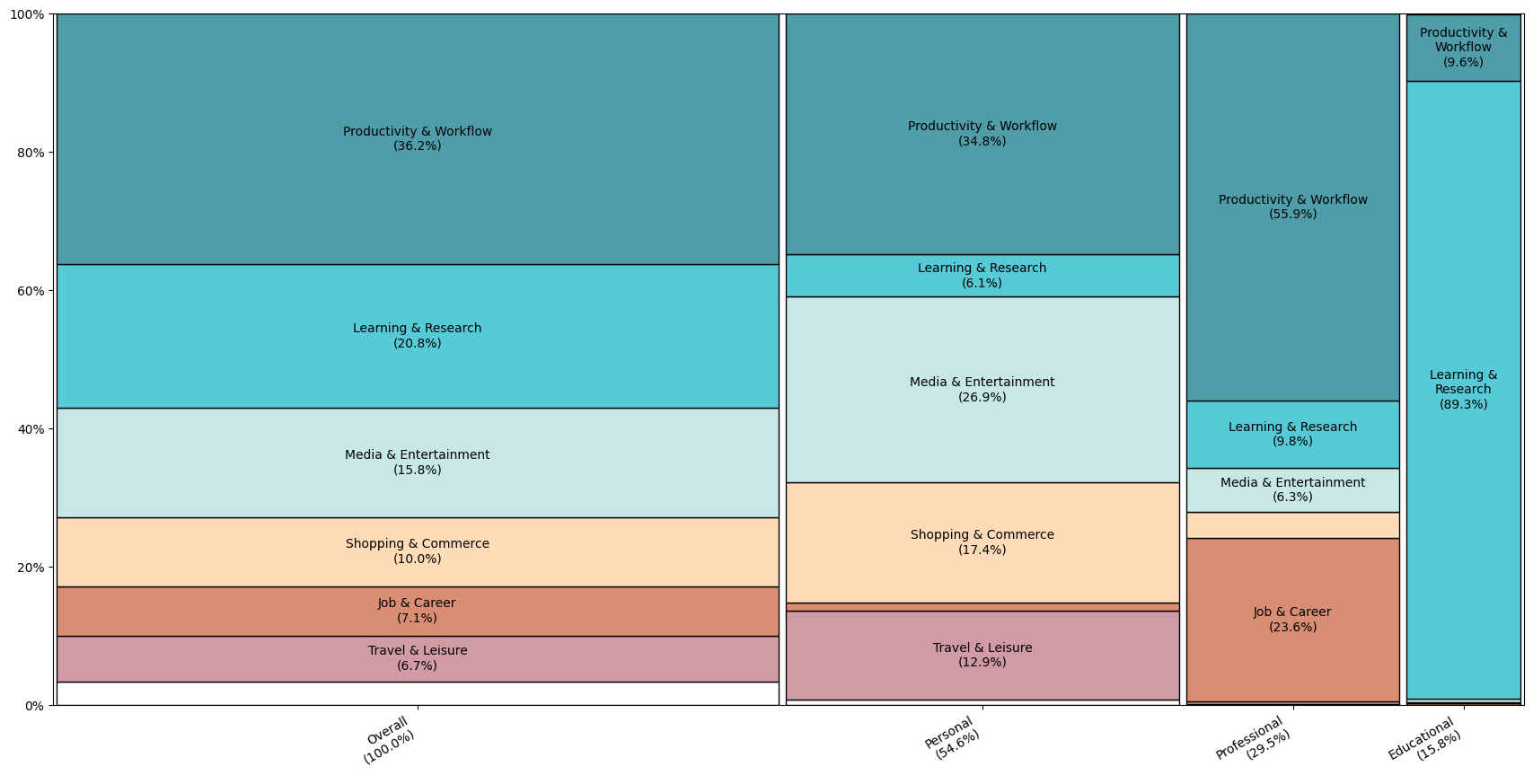}
    \caption{Topic Distribution by Usage Context}
    \begin{minipage}[t]{\textwidth}
      \raggedright
      \footnotesize
      Note: The plot shows the percentage shares of topics within each usage context. The ``Other'' category is removed from contexts. Bar width is proportional to context percentage, and box height within each bar is proportional to topic percentage. The labels for topics that account for less than 5\% within a context are suppressed. Context shares are shown in the x-axis labels. The same topic is shown in the same color across contexts. The topic percentage overall is shown as a baseline for comparison. The topics within each context are sorted in the same order as the overall for easier comparison across contexts. The topics in the overall category are sorted by topic percentage, from highest to lowest.
    \end{minipage}
    \label{fig:use case-context-topic}
\end{figure}

We show the distribution of topics by usage context in Figure \ref{fig:use case-context-topic}. For personal use, productivity and media together account for 62\% of all agentic queries. For professional use, 80\% of agentic queries are productivity- and career-related. Educational usage is dominated by learning, comprising 89\% of agentic queries. We also show the distribution of subtopic, task, and environment by usage context in Tables \ref{tab:use case-context-subtopic}, \ref{tab:use case-context-task}, and \ref{tab:use case-context-env} in Appendix \ref{figures and tables}. The top subtopics for personal, professional, and educational use are goods shopping, document editing, and courses, respectively. The top environments for personal, professional, and educational use are emails, linkedin.com, and docs.google.com, respectively.

\section{Discussion}\label{discussion}
Our paper provides the first systematic evidence on the adoption, usage intensity, and use cases of general-purpose AI agents, based on large-scale behavioral data from Comet by Perplexity. Our findings reveal substantial differences in the propensity to adopt and use the agent across user segments. Earlier adopters, users in countries with higher GDP per capita and higher average years of education, and individuals working in more digital or knowledge-intensive fields—such as digital technology, academia, finance, marketing, and entrepreneurship—tend to adopt and use the agent more actively. Agent use cases span a broad range of categories. The two largest topics—productivity and learning—together comprise 57\% of all agentic queries. The two largest subtopics—courses and goods shopping—together account for 22\% of all agentic queries. The top 10 out of 90 tasks represent 55\% of all agentic queries. We also document heterogeneity in use cases across occupation clusters, reflecting the degree to which they align with each occupation's task composition. Topics such as productivity, learning, and career exhibit higher stickiness, as users are more likely to make consecutive queries within these categories. Over time, users also shift toward more cognitively oriented tasks. In addition, the environments in which agentic queries are made show significant variation in concentration across topics and subtopics.

Although our paper is primarily descriptive and does not make normative claims or directly examine downstream impacts, its methods and findings offer valuable implications for researchers, businesses, policymakers, and educators. For researchers, we contribute to a nascent but rapidly expanding literature on the adoption and usage of LLMs and AI agents, and our agentic taxonomy provides a structure for future analysis to build on and extend. For firms developing AI agents, our results offer guidance on target user segments and high-frequency use cases. For businesses that provide the environments in which agents operate, our findings suggest opportunities to streamline interfaces to better serve users interacting with AI agents. For both policymakers and educators, a central concern is that uneven adoption and usage of AI agents could exacerbate existing productivity and learning disparities. Consequently, equipping citizens and students with the skills to leverage AI agents effectively and preparing them for a near future in which such agents are embedded in work and everyday life will become increasingly important. 

We note a few important caveats of our dataset. First, because Comet is a new product, our sample primarily reflects early adopters, who may skew toward more tech-savvy users. We characterize these early adopters using an internal survey in Appendix~\ref{survey}. Relatedly, given the short time span of our data, we do not systematically investigate changes in usage patterns over time, and any longitudinal results should be interpreted within this context. Second, the classification of an agentic query depends on internal query understanding modules that trigger the agent based on predicted query intent. These intent predictors show high prediction accuracy in internal validation studies; nonetheless, the data may include both false positives (when a non-agentic intent triggers the agent) and false negatives (when an agentic intent does not trigger the agent).\footnote{Two sets of classifiers determine agent activation: one for browser control and one for specific apps. The browser control classifier is a supervised model trained on user queries labeled as either showing or not showing agentic intent. Five features are used for prediction: the current query, the currently viewed page, the user's previous queries in the same conversation, enabled connectors, and the number of attachments. The classifier achieves an ROC-AUC of 0.95 and both precision and recall of 0.90 at the optimal threshold. Each connector additionally has its own classifier, following a similar process.} Similarly, the classification into our agentic taxonomy, usage context, and occupation clusters also contains noise. Third, although AI agents' autonomy and task horizons continue to expand, our results should not be interpreted as suggesting any particular balance between automation and augmentation in use cases. For instance, an agentic session may appear to be automation, but users may break a task into smaller pieces and delegate only some subtasks to the agent, which is closer to an augmentation case. A comprehensive treatment on such a topic would require having a complete picture of how users manage their workflows outside of Comet.

There are several natural extensions to this study that we aim to pursue. First, with the expansion of Comet to mobile devices and other environments, it will be valuable to document cross-platform differences in how users interact with AI agents.\footnote{The Android version of Comet was launched to everyone worldwide on November 20, 2025, with the iOS version scheduled to be released in December 2025. \url{https://www.perplexity.ai/hub/blog/comet-for-android-is-here}} In particular, whereas agentic queries on desktops are predominantly text-based, the voice-to-voice mode on mobile may offer a more natural interface. Second, while our sample does not capture enterprise users directly, the substantial share of professionally-oriented agentic queries suggests the need for complementary research on related topics in organizational settings. Third, the adoption and usage of AI agents is closely tied to their performance across tasks; we plan to investigate agent evaluation, common failure modes, and strategies for improvement. Fourth, identifying which tasks are best suited for delegation to the agent and designing optimal human-agent collaborative workflows are also important questions. For example, tasks that users can easily complete manually may not warrant delegation. High-stakes or irreversible tasks might require exceptionally reliable agent performance, a high degree of user trust, and increased human supervision. Fifth, in addition to awareness and performance, another key barrier to adoption and use is measuring and substantiating value and impact; we seek to quantify the economic value users derive from agent use, an essential dimension of the downstream impact of AI agents. 

General-purpose AI agents represent one of the most consequential technological advancements of our time. Understanding their real-world adoption and usage with large-scale behavioral data has become both urgent and essential for informing their development and deployment. We hope this work catalyzes further investigation in this rapidly evolving domain.

\clearpage 
\bibliographystyle{abbrvnat}
\bibliography{references}  

\clearpage 
\appendix \label{appendix}
\renewcommand*{\appendixpagename}{\large Appendices} 
\appendixpage

\section{Figures and Tables} \label{figures and tables}

\subsection{Adoption and Usage Intensity}

\vspace*{\fill}
\begin{figure}[H]
  \centering
  \begin{subfigure}[b]{0.8\textwidth}
    \includegraphics[width=\linewidth]{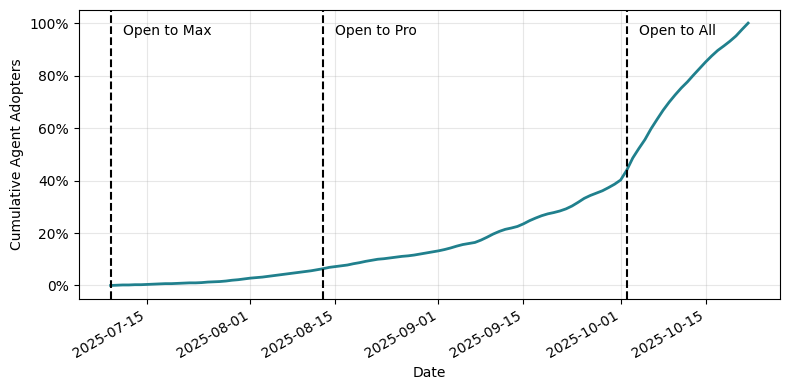}
    \caption{Cumulative Agent Adopter}
  \end{subfigure}
  
  \vspace{0.5cm} 

  \begin{subfigure}[b]{0.8\textwidth}
    \includegraphics[width=\linewidth]{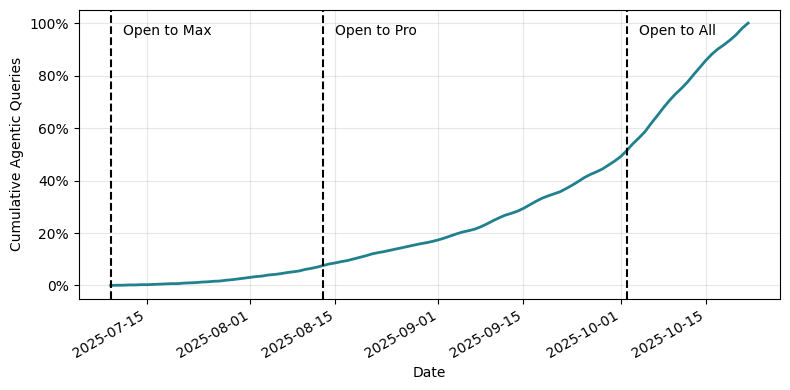}    
    \caption{Cumulative Agentic Query}

  \end{subfigure}
  \caption{Cumulative Agent Adopter and Agentic Query}

    \begin{minipage}[t]{\textwidth}
      \raggedright
       \scriptsize
      Note: The plots show the cumulative number of agent adopters and agentic queries. The exact numbers on the y-axis are masked, and the percentages show the relative magnitude of adopters and queries relative to the end date as the baseline. Adopter and query numbers grow steadily over time with a noticeable jump when Comet became generally available. The three dashed vertical lines mark the dates for the change in access. July 9: launch date and open to Max subscribers and selected users on a waitlist. August 13: extended access to Pro subscribers. October 2: extended access to everyone worldwide. 
    \end{minipage}

  \label{fig:adoption-usage}
\end{figure}

\clearpage
\vspace*{\fill}
  
\begin{table}[H]
  \centering
  \rotatebox{90}{%
\begin{threeparttable}
\centering
\footnotesize

\begin{tabular}{llccc}
\toprule
\multicolumn{5}{c}{Agent Adoption by Occupation Subcluster: The Top 10 Subclusters by Agent Adoption Share} \\
\midrule
\textbf{Cluster} & \textbf{Subcluster} & \textbf{User Share (\%)} & \textbf{Agent Adopter Share (\%)} & \textbf{AAR} \\
\midrule
Digital Technology              & Software Development \& Engineering       & 13.0 & 13.9 & 1.07 \\
Marketing \& Sales              & Digital Marketing \& Social Media         & 3.9  & 4.7  & 1.21 \\
Digital Technology              & IT Support \& Infrastructure              & 4.0  & 4.3  & 1.08 \\
Arts, Entertainment, \& Design  & Design \& Digital Arts                    & 5.0  & 4.3  & 0.86 \\
Financial Services              & Financial Planning \& Analysis            & 4.1  & 4.3  & 1.05 \\
Digital Technology              & Data Science \& AI                       & 3.9  & 4.0  & 1.03 \\
Education                       & Teaching \& Instruction                   & 4.9  & 4.0  & 0.82 \\
Financial Services              & Financial Strategy \& Investments         & 3.8  & 3.9  & 1.03 \\
Management \& Entrepreneurship  & Business Information Management           & 3.2  & 3.8  & 1.19 \\
Management \& Entrepreneurship  & Leadership \& Operations                  & 2.6  & 3.1  & 1.19 \\
\midrule
\multicolumn{5}{c}{Agent Adoption by Occupation Subcluster: The Top 10 Subclusters by AAR} \\
\midrule
\textbf{Cluster} & \textbf{Subcluster} & \textbf{User Share (\%)} & \textbf{Agent Adopter Share (\%)} & \textbf{AAR} \\
\midrule
Marketing \& Sales              & Business Development \& Sales             & 1.2  & 1.6  & 1.33 \\
Marketing \& Sales              & Digital Marketing \& Social Media         & 3.9  & 4.7  & 1.21 \\
Management \& Entrepreneurship  & Business Information Management           & 3.2  & 3.8  & 1.19 \\
Management \& Entrepreneurship  & Leadership \& Operations                  & 2.6  & 3.1  & 1.19 \\
Marketing \& Sales              & Market Research, Analytics, \& Ethics     & 1.1  & 1.3  & 1.18 \\
Management \& Entrepreneurship  & Strategy \& Consulting                    & 2.2  & 2.5  & 1.14 \\
Supply Chain \& Transportation  & Planning \& Logistics                     & 1.1  & 1.2  & 1.09 \\
Digital Technology              & IT Support \& Infrastructure              & 4.0  & 4.3  & 1.08 \\
Digital Technology              & Software Development \& Engineering       & 13.0 & 13.9 & 1.07 \\
Marketing \& Sales              & Brand Management \& Strategy              & 1.4  & 1.5  & 1.07 \\
\bottomrule
\end{tabular}

\begin{tablenotes}
 \scriptsize
\item Note: The tables show the top 10 O*NET occupation subclusters by adoption. The ``Other'' category is removed. When a user appears in multiple subclusters, their data is used in all relevant subclusters. User share is the number of users in each subcluster divided by the total users. Agent adopter share is the number of adopters in each subcluster, divided by the total number of adopters. AAR (Agent Adoption Ratio) is the ratio between agent adopter share and user share. AAR greater (less) than 1 indicates that a subcluster is over-represented (under-represented) in agent adoption relative to their user base. AAR rank is among the subclusters with a user share over 1\%.
\end{tablenotes}
\caption{Agent Adoption by Occupation Subcluster}
\label{tab:adoption-occupation-subcluster}
\end{threeparttable}
}
\end{table}
\vspace*{\fill}

\clearpage
\vspace*{\fill}

\begin{table}[H]

  \rotatebox{90}{%

\begin{threeparttable}
\footnotesize

\begin{tabular}{llccc}
\toprule
\multicolumn{5}{c}{Agentic Query by Occupation Subcluster: The Top 10 Subclusters by Agentic Query Share} \\
\midrule
\textbf{Cluster} & \textbf{Subcluster} & \textbf{User Share (\%)} & \textbf{Agentic Query Share (\%)} & \textbf{AUR} \\
\midrule
Digital Technology              & Software Development \& Engineering       & 13.0 & 15.4 & 1.18 \\
Marketing \& Sales              & Digital Marketing \& Social Media         & 3.9  & 5.9  & 1.51 \\
Digital Technology              & IT Support \& Infrastructure              & 4.0  & 5.3  & 1.32 \\
Management \& Entrepreneurship  & Business Information Management           & 3.2  & 4.7  & 1.47 \\
Financial Services              & Financial Planning \& Analysis            & 4.1  & 3.9  & 1.05 \\
Digital Technology              & Data Science \& AI                        & 3.9  & 3.9  & 1.00 \\
Management \& Entrepreneurship  & Leadership \& Operations                  & 2.6  & 3.9  & 1.50 \\
Arts, Entertainment, \& Design  & Design \& Digital Arts                    & 5.0  & 3.8  & 0.76 \\
Education                       & Teaching \& Instruction                   & 4.9  & 3.6  & 0.73 \\
Financial Services              & Financial Strategy \& Investments         & 3.8  & 3.9  & 1.03 \\
\midrule
\multicolumn{5}{c}{Agentic Query by Occupation Subcluster: The Top 10 Subclusters by AUR} \\
\midrule
\textbf{Cluster} & \textbf{Subcluster} & \textbf{User Share (\%)} & \textbf{Agentic Query Share (\%)} & \textbf{AUR} \\
\midrule
Marketing \& Sales              & Business Development \& Sales             & 1.2  & 2.1  & 1.75 \\
Marketing \& Sales              & Digital Marketing \& Social Media         & 3.9  & 5.9  & 1.51 \\
Management \& Entrepreneurship  & Leadership \& Operations                  & 2.6  & 3.9  & 1.50 \\
Management \& Entrepreneurship  & Business Information Management           & 3.2  & 4.7  & 1.47 \\
Marketing \& Sales              & Brand Management \& Strategy              & 1.4  & 1.9  & 1.36 \\
Management \& Entrepreneurship  & Strategy \& Consulting                    & 2.2  & 3.0  & 1.36 \\
Digital Technology              & IT Support \& Infrastructure              & 4.0  & 5.3  & 1.32 \\
Marketing \& Sales              & Market Research, Analytics, \& Ethics     & 1.1  & 1.4  & 1.27 \\
Digital Technology              & Software Development \& Engineering       & 13.0 & 15.4 & 1.18 \\
Supply Chain \& Transportation  & Planning \& Logistics                     & 1.1  & 1.1  & 1.00 \\
\bottomrule
\end{tabular}

\begin{tablenotes}
\scriptsize
\item Note: The tables show the top 10 O*NET occupation subclusters by usage intensity. The ``Other'' category is removed. When a user appears in multiple subclusters, their data is used in all relevant subclusters. User share is the number of users in each subcluster divided by the total users. Agent query share is the number of agentic queries in each subcluster divided by the total agentic queries. AUR (Agent Usage Ratio) is the ratio between agentic query share and user share. AUR greater (less) than 1 indicates that a subcluster is over-represented (under-represented) in agent usage relative to their user base. AUR rank is among the subclusters with a user share over 1\%. 
\end{tablenotes}
\caption{Agentic Query by Occupation Subcluster}
\label{tab:usage-occupation-subcluster}
\end{threeparttable}
}
\end{table}
\vspace*{\fill}

\clearpage
\subsection{Use Cases}
\subsubsection{Topics and Subtopics}
\vspace*{\fill}
\begin{table}[H]
\footnotesize
\centering
\begin{threeparttable}
\caption{Topic and Subtopic Distribution}
\label{tab:use case-topic-subtopic}
\begin{tabular}{lclcc}
\hline
\textbf{Topic} & \textbf{Topic (\%)} & \textbf{Subtopic} & \textbf{Subtopic (\%)} & \textbf{Overall (\%)} \\
\hline
\multirow{10}{*}{Productivity \& Workflow} 
  & \multirow{10}{*}{36.2} & Document \& Form Editing      & 21.5 & 7.78 \\
  &                  & Account Management          & 20.5 & 7.43 \\
  &                  & Email Management  & 15.8 & 5.73 \\
  &                  & Spreadsheet \& Data Editing   & 11.1 & 4.01 \\
  &                  & Computer Programming          & 10.3 & 3.73 \\
  &                  & Investments \& Banking        & 6.2  & 2.25 \\
  &                  & Multimedia Editing            & 6.1  & 2.22 \\
  &                  & Project Management            & 5.1  & 1.85 \\
  &                  & Calendar Management          & 2.5  & 0.91 \\
  &                  & Other                         & 0.8  & 0.30 \\
\cline{1-5}
\multirow{3}{*}{Learning \& Research} 
  & \multirow{3}{*}{20.8} & Courses        & 61.9 & 12.86 \\
  &                  & Research          & 37.9 & 7.88 \\
  &                  & Other                         & 0.2  & 0.04 \\
\cline{1-5}
\multirow{7}{*}{Media \& Entertainment}
  & \multirow{7}{*}{15.8} & Social Media \& Messaging     & 42.4 & 6.69 \\
  &                  & Movies, TV, \& Videos         & 20.1 & 3.17 \\
  &                  & Online Games                  & 19.6 & 3.08 \\
  &                  & Music \& Podcasts             & 10.7 & 1.68 \\
  &                  & News                          & 3.8  & 0.59 \\
  &                  & Sports                        & 2.7  & 0.42 \\
  &                  & Other                         & 0.8  & 0.13 \\
\cline{1-5}
\multirow{3}{*}{Shopping \& Commerce}
  & \multirow{3}{*}{10.0} & Goods                         & 89.0 & 8.94 \\
  &                  & Services                      & 10.3 & 1.03 \\
  &                  & Other                         & 0.7  & 0.07 \\
\cline{1-5}
\multirow{3}{*}{Job \& Career}
  & \multirow{3}{*}{7.1} & Professional Networking       & 50.1 & 3.56 \\
  &                  & Job Search \& Application     & 49.5 & 3.52 \\
  &                  & Other                         & 0.4  & 0.03 \\
\cline{1-5}
\multirow{5}{*}{Travel \& Leisure}
  & \multirow{5}{*}{6.7} & Flights \& Transportation     & 40.7 & 2.73 \\
  &                  & Lodging                       & 27.3 & 1.83 \\
  &                  & Trip Itineraries              & 24.8 & 1.66 \\
  &                  & Restaurants                   & 5.7  & 0.38 \\
  &                  & Other                         & 1.5  & 0.10 \\
\cline{1-5}
Other & 3.4 & Other & 100.0 & 3.42 \\
\hline
\end{tabular}
\begin{tablenotes}
\footnotesize
\item Note: The table shows the distribution of topics and subtopics. The topic percentage ($P(\text{Topic})$) is the topic's share among all agentic queries. Subtopic percentage ($P(\text{Subtopic | Topic})$) is the percentage of a subtopic within a topic. Overall percentage ($P(\text{Subtopic}) = P(\text{Topic, Subtopic}) = P(\text{Topic}) \times P(\text{Subtopic | Topic})$) is the subtopic share among all agentic queries. Percentages may not sum to 100\% due to rounding.
\end{tablenotes}
\end{threeparttable}
\end{table}
\vspace*{\fill}

\clearpage
\clearpage
\vspace*{\fill}
\begin{figure}[H]
  \centering
     \includegraphics[width=.9\linewidth]{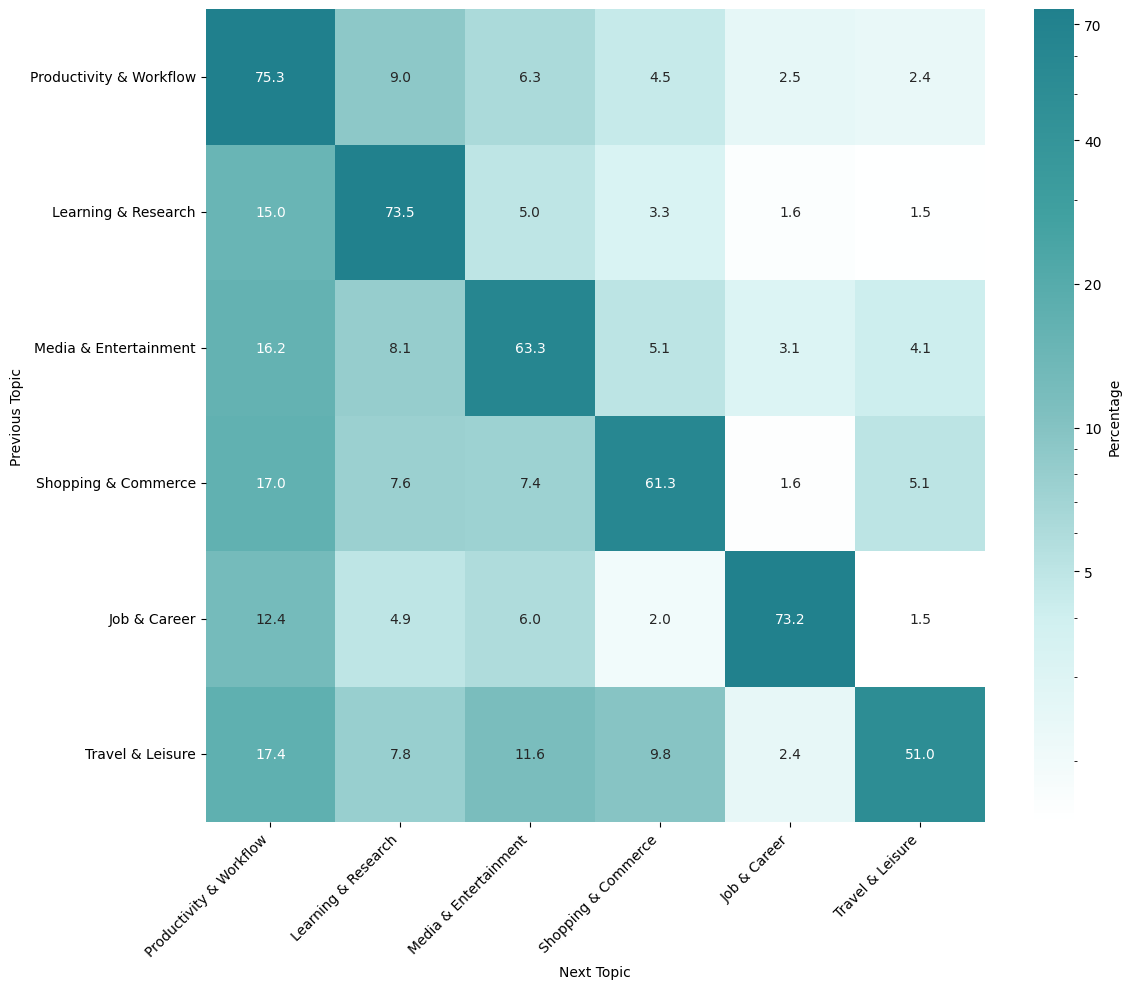}
    \caption{Topic Transition Matrix: Previous vs. Next Agentic Query}
      \label{fig:use case-topic transition}
    \begin{minipage}[t]{\textwidth}
      \raggedright
      \footnotesize
       Note: The plot shows the transition matrix from the previous query to the following query, aggregated from the user level. Most query topics transition into themselves (the off-diagonal). Other than themselves, topics are most likely to transition into Productivity \& Workflow (the first column). Productivity \& Workflow, Learning \& Research, and Job \& Career are the most sticky with the highest self-transition probabilities. Whereas Travel \& Leisure is the least sticky, and Media \& Entertainment and Shopping \& Commerce are in between. The steady state probability distribution based on this transition matrix and the observed share are closely matched—39\% vs 37\% respectively for Productivity \& Workflow, 24\% vs 22\% for Learning \& Research, 16\% vs. 16\% for Media \& Entertainment, 10\% vs. 10\% for Shopping \& Commerce, 7\% vs 7\% for Job \& Career, and 7\% vs 7\% for Travel \& Leisure. Note that to be aligned with the transition matrix, the observed shares used here do not include the ``Other'' category in Table \ref{tab:use case-topic-subtopic}; the shares used here are based on the topic shares in that table divided by 0.96.
    \end{minipage}
\end{figure}
\vspace*{\fill}

\clearpage
\vspace*{\fill}
\begin{figure}[H]
  \centering
    \includegraphics[width=.8\linewidth]{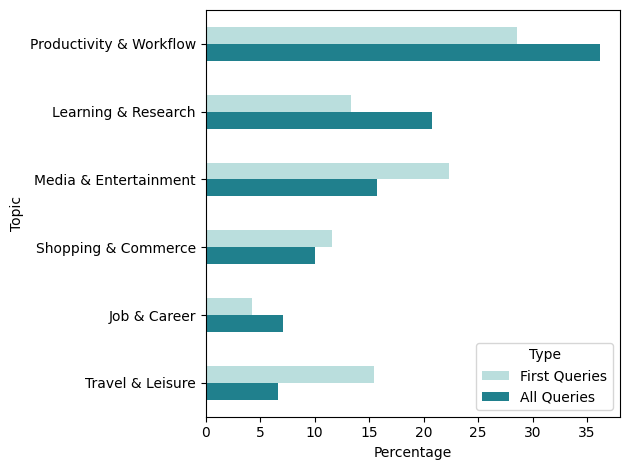}
    \caption{Topic Distribution by First vs. All Agentic Queries}
      \label{fig:use case-first vs all}
    \begin{minipage}[t]{\textwidth}
      \raggedright
      \footnotesize
      Note: The plot shows the distribution of topics among users' first agentic queries versus all agentic queries. Over time, the query shares are shifting from Travel \& Leisure, and Media \& Entertainment to Productivity \& Workflow, Learning \& Research, and Job \& Career. Shopping \& Commerce share stays relatively stable. 
    \end{minipage}
\end{figure}
\vspace*{\fill}
\clearpage

\clearpage
\subsubsection{Tasks}

\vspace*{\fill}
\begin{table}[H]
\centering
\footnotesize
\begin{threeparttable}
\begin{tabular}{
    >{\centering\arraybackslash}m{3.2cm}  
    >{\centering\arraybackslash}m{3.6cm}  
    l                                     
    c                                     
    c                                     
}
\toprule
\textbf{Topic} & \textbf{Subtopic} & \textbf{Task} & \textbf{Task (\%)} & \textbf{Overall (\%)} \\
\midrule

\multirow{30}{*}{\centering\shortstack{Productivity \& Workflow \\ (36.2)}}

  & \multirow{3}{*}{\centering\shortstack{Document \& Form Editing \\ (21.5)}}
      & Create/edit documents/forms & 84.6 & 6.58 \\
  &   & Summarize/analyze documents/forms & 24.1 & 1.87 \\
  &   & Search/filter documents/forms & 11.4 & 0.89 \\[2pt]
\cline{2-5}

  & \multirow{4}{*}{\centering\shortstack{Account Management \\ (20.5)}}
      & Manage settings/profiles & 58.3 & 4.33 \\
  &   & Register/log in to accounts & 31.2 & 2.32 \\
  &   & Summarize/analyze account information & 24.0 & 1.79 \\
  &   & Manage files & 15.8 & 1.18 \\[2pt]
\cline{2-5}

  & \multirow{5}{*}{\centering\shortstack{Email Management \\ (15.8)}}
      & Search/filter emails & 49.1 & 2.81 \\
  &   & Create/edit emails & 32.8 & 1.88 \\
  &   & Delete/unsubscribe emails & 30.6 & 1.75 \\
  &   & Summarize/analyze emails & 22.8 & 1.31 \\
  &   & Send emails & 9.6 & 0.55 \\[2pt]
\cline{2-5}

  & \multirow{3}{*}{\centering\shortstack{Spreadsheet \& Data Editing \\ (11.1)}}
      & Create/edit spreadsheets/data & 72.5 & 2.91 \\
  &   & Summarize/analyze spreadsheets/data & 38.7 & 1.55 \\
  &   & Search/filter spreadsheets/data & 27.5 & 1.10 \\[2pt]
\cline{2-5}

  & \multirow{3}{*}{\centering\shortstack{Computer Programming \\ (10.3)}}
      & Create/edit code & 63.8 & 2.38 \\
  &   & Summarize/analyze code & 48.5 & 1.81 \\
  &   & Execute code & 20.3 & 0.76 \\[2pt]
\cline{2-5}

  & \multirow{4}{*}{\centering\shortstack{Investments \& Banking \\ (6.2)}}
      & Summarize/analyze investment information & 75.0 & 1.69 \\
  &   & Search/filter stocks & 28.8 & 0.65 \\
  &   & Summarize/analyze banking information & 11.7 & 0.26 \\
  &   & Buy/sell stocks & 7.8 & 0.18 \\[2pt]
\cline{2-5}

  & \multirow{3}{*}{\centering\shortstack{Multimedia Editing \\ (6.1)}}
      & Create/edit multimedia & 81.0 & 1.80 \\
  &   & Summarize/analyze multimedia & 23.6 & 0.52 \\
  &   & Search/filter multimedia & 15.3 & 0.34 \\[2pt]
\cline{2-5}

  & \multirow{2}{*}{\centering\shortstack{Project Management \\ (5.1)}}
      & Create/edit projects & 64.6 & 1.19 \\
  &   & Summarize/analyze project information & 48.0 & 0.89 \\[2pt]
\cline{2-5}

  & \multirow{3}{*}{\centering\shortstack{Calendar Management \\ (2.5)}}
      & Create/edit events & 71.0 & 0.64 \\
  &   & Search/filter events & 24.6 & 0.22 \\
  &   & Summarize/analyze events & 22.0 & 0.20 \\

\bottomrule
\end{tabular}
\caption{Task Distribution for Productivity \& Workflow}
\label{tab:use case-productivity-task}
\begin{tablenotes}
\scriptsize
    \item Note: The table shows all tasks under Productivity \& Workflow with a share of more than 5\% within the subtopic. The share of the topic among all agentic queries and the share of the subtopic within a topic are shown in parentheses. Task percentage ($P(\text{Task | Topic, Subtopic})$) is the task share within the subtopic. The overall percentage is the task share among all agentic queries. $P(\text{Task Overall}) = P(\text{Topic, Subtopic, Task}) = P(\text{Topic}) \times P(\text{Subtopic | Topic}) \times P(\text{Task | Topic, Subtopic}) $). Note that because task percentage measures the fraction of queries in which a task is present and a query might contain multiple tasks, the task percentages under each subtopic do not sum to 100.  
\end{tablenotes}
\end{threeparttable}
\end{table}
\vspace*{\fill}
\clearpage
\vspace*{\fill}
\begin{table}[H]
\centering
\begin{threeparttable}
\footnotesize
\begin{tabular}{
    >{\centering\arraybackslash}m{3.4cm}  
    >{\centering\arraybackslash}m{3.8cm}  
    l                                     
    c                                     
    c                                     
}
\toprule
\textbf{Topic} & \textbf{Subtopic} & \textbf{Task} & \textbf{Task (\%)} & \textbf{Overall (\%)} \\
\midrule

\multirow{5}{*}{\centering\shortstack{Learning \& Research \\ (20.8)}}

  & \multirow{3}{*}{\centering\shortstack{Courses \\ (61.9)}}
      & Assist exercises & 73.2 & 9.41 \\
  &   & Summarize/analyze course materials & 28.7 & 3.69 \\
  &   & Navigate courses & 25.6 & 3.29 \\[2pt]
\cline{2-5}

  & \multirow{2}{*}{\centering\shortstack{Research \\ (37.9)}}
      & Summarize/analyze research information & 85.2 & 6.71 \\
  &   & Search/filter research information & 75.6 & 5.95 \\

\bottomrule
\end{tabular}
\caption{Task Distribution for Learning \& Research}
\label{tab:use case-learning-task}
\begin{tablenotes}
\scriptsize
    \item Note: The table shows all tasks under Learning \& Research with a share of more than 5\% within the subtopic. The share of the topic among all agentic queries and the share of the subtopic within a topic are shown in parentheses. Task percentage ($P(\text{Task | Topic, Subtopic})$) is the task share within the subtopic. The overall percentage is the task share among all agentic queries. $P(\text{Task Overall}) = P(\text{Topic, Subtopic, Task}) = P(\text{Topic}) \times P(\text{Subtopic | Topic}) \times P(\text{Task | Topic, Subtopic}) $). Note that because task percentage measures the fraction of queries in which a task is present and a query might contain multiple tasks, the task percentages under each subtopic do not sum to 100.
\end{tablenotes}
\end{threeparttable}
\end{table}
\vspace*{\fill}

\clearpage
\vspace*{\fill}
\begin{table}[H]
\centering
\begin{threeparttable}
\footnotesize
\begin{tabular}{
    >{\centering\arraybackslash}m{3.0cm}   
    >{\centering\arraybackslash}m{3.6cm}   
    l                                      
    c                 
    c                 
}
\toprule
\textbf{Topic} & \textbf{Subtopic} & \textbf{Task} & \textbf{Task (\%)} & \textbf{Overall (\%)} \\
\midrule

\multirow{20}{*}{\centering\shortstack{Media \& Entertainment \\ (15.8)}}

  & \multirow{5}{*}{\centering\shortstack{Social Media \& Messaging \\ (42.4)}}
      & Search/filter social media posts/messages & 49.5 & 3.31 \\
  &   & Summarize/analyze social media posts/messages & 35.3 & 2.36 \\
  &   & Create social media posts/messages & 34.0 & 2.28 \\
  &   & Engage with social media posts/messages & 29.3 & 1.96 \\
  &   & Send social media/text messages & 20.7 & 1.39 \\[2pt]
\cline{2-5}

  & \multirow{5}{*}{\centering\shortstack{Movies, TV, \& Videos \\ (20.1)}}
      & Search/filter videos & 48.4 & 1.53 \\
  &   & Summarize/analyze videos & 43.8 & 1.39 \\
  &   & Play videos & 27.1 & 0.86 \\
  &   & Navigate within videos & 18.4 & 0.58 \\
  &   & Manage playlists & 6.2 & 0.20 \\[2pt]
\cline{2-5}

  & \multirow{3}{*}{\centering\shortstack{Online Games \\ (19.6)}}
      & Play online games & 76.8 & 2.37 \\
  &   & Summarize/analyze online game information & 30.7 & 0.95 \\
  &   & Search/filter online games & 14.5 & 0.45 \\[2pt]
\cline{2-5}

  & \multirow{4}{*}{\centering\shortstack{Music \& Podcasts \\ (10.7)}}
      & Search/filter music/podcasts & 75.4 & 1.27 \\
  &   & Play music/podcasts & 61.2 & 1.03 \\
  &   & Manage playlists & 25.5 & 0.43 \\
  &   & Summarize/analyze music/podcasts & 9.7 & 0.16 \\[2pt]
\cline{2-5}

  & \multirow{2}{*}{\centering\shortstack{News \\ (3.8)}}
      & Search/filter news & 70.7 & 0.42 \\
  &   & Summarize/analyze news & 56.6 & 0.34 \\[2pt]
\cline{2-5}

  & \multirow{2}{*}{\centering\shortstack{Sports \\ (2.7)}}
      & Summarize/analyze match/player information & 77.0 & 0.32 \\
  &   & Search/filter match/player information & 67.7 & 0.28 \\

\bottomrule
\end{tabular}
\caption{Task Distribution for Media \& Entertainment}
\label{tab:use case-media-task}
\begin{tablenotes}
\scriptsize
    \item Note: The table shows all tasks under Media \& Entertainment with a share of more than 5\% within the subtopic. The share of the topic among all agentic queries and the share of the subtopic within a topic are shown in parentheses. Task percentage ($P(\text{Task | Topic, Subtopic})$) is the task share within the subtopic. The overall percentage is the task share among all agentic queries. $P(\text{Task Overall}) = P(\text{Topic, Subtopic, Task}) = P(\text{Topic}) \times P(\text{Subtopic | Topic}) \times P(\text{Task | Topic, Subtopic}) $). Note that because task percentage measures the fraction of queries in which a task is present and a query might contain multiple tasks, the task percentages under each subtopic do not sum to 100. 
\end{tablenotes}
\end{threeparttable}
\end{table}
\vspace*{\fill}
\clearpage
\vspace*{\fill}
\begin{table}[H]
\centering
\begin{threeparttable}
\footnotesize
\begin{tabular}{
    >{\centering\arraybackslash}m{3.2cm}   
    >{\centering\arraybackslash}m{3.8cm}   
    l                                      
    c                                      
    c                                      
}
\toprule
\textbf{Topic} & \textbf{Subtopic} & \textbf{Task} & \textbf{Task (\%)} & \textbf{Overall (\%)} \\
\midrule

\multirow{11}{*}{\centering\shortstack{Shopping \& Commerce \\ (10.0)}}

  & \multirow{4}{*}{\centering\shortstack{Goods \\ (89.0)}}
      & Search/filter products & 71.9 & 6.43 \\
  &   & Summarize/analyze product information & 57.9 & 5.18 \\
  &   & Add products to cart & 19.8 & 1.77 \\
  &   & Search discounts & 10.2 & 0.92 \\[2pt]
\cline{2-5}

  & \multirow{6}{*}{\centering\shortstack{Services \\ (10.3)}}
      & Search/filter products & 54.5 & 0.56 \\
  &   & Summarize/analyze product information & 45.1 & 0.46 \\
  &   & Make product purchase & 20.2 & 0.21 \\
  &   & Search discounts & 12.5 & 0.13 \\
  &   & Add products to cart & 7.6 & 0.08 \\
  &   & Manage orders & 7.0 & 0.07 \\

\bottomrule
\end{tabular}
\caption{Task Distribution for Shopping \& Commerce}
\label{tab:use case-shopping-task}
\begin{tablenotes}
\scriptsize
    \item Note: The table shows all tasks under Shopping \& Commerce with a share of more than 5\% within the subtopic. The share of the topic among all agentic queries and the share of the subtopic within a topic are shown in parentheses. Task percentage ($P(\text{Task | Topic, Subtopic})$) is the task share within the subtopic. The overall percentage is the task share among all agentic queries. $P(\text{Task Overall}) = P(\text{Topic, Subtopic, Task}) = P(\text{Topic}) \times P(\text{Subtopic | Topic}) \times P(\text{Task | Topic, Subtopic}) $). Note that because task percentage measures the fraction of queries in which a task is present and a query might contain multiple tasks, the task percentages under each subtopic do not sum to 100.
\end{tablenotes}
\end{threeparttable}
\end{table}
\vspace*{\fill}

\clearpage
\vspace*{\fill}
\begin{table}[H]
\centering
\begin{threeparttable}
\footnotesize
\begin{tabular}{
    >{\centering\arraybackslash}m{3.2cm}   
    >{\centering\arraybackslash}m{3.8cm}   
    l                                      
    c                                      
    c                                      
}
\toprule
\textbf{Topic} & \textbf{Subtopic} & \textbf{Task} & \textbf{Task (\%)} & \textbf{Overall (\%)} \\
\midrule

\multirow{7}{*}{\centering\shortstack{Job \& Career \\ (7.1)}}

  & \multirow{3}{*}{\centering\shortstack{Job Search \& Application \\ (50.1)}}
      & Complete applications & 65.7 & 2.31 \\
  &   & Search/filter jobs & 57.1 & 2.01 \\
  &   & Summarize/analyze job descriptions & 26.3 & 0.93 \\[2pt]
\cline{2-5}

  & \multirow{4}{*}{\centering\shortstack{Professional Networking \\ (49.5)}}
      & Search/filter professional profiles & 53.8 & 1.92 \\
  &   & Summarize/analyze professional profiles & 39.9 & 1.42 \\
  &   & Send professional connection requests/messages & 31.4 & 1.12 \\
  &   & Engage with professional profiles/posts & 24.0 & 0.85 \\

\bottomrule
\end{tabular}
\caption{Task Distribution for Job \& Career}
\label{tab:use case-job-task}
\begin{tablenotes}
\scriptsize
    \item Note: The table shows all tasks under Job \& Career with a share of more than 5\% within the subtopic. The share of the topic among all agentic queries and the share of the subtopic within a topic are shown in parentheses. Task percentage ($P(\text{Task | Topic, Subtopic})$) is the task share within the subtopic. The overall percentage is the task share among all agentic queries. $P(\text{Task Overall}) = P(\text{Topic, Subtopic, Task}) = P(\text{Topic}) \times P(\text{Subtopic | Topic}) \times P(\text{Task | Topic, Subtopic}) $). Note that because task percentage measures the fraction of queries in which a task is present and a query might contain multiple tasks, the task percentages under each subtopic do not sum to 100.
\end{tablenotes}
\end{threeparttable}
\end{table}
\vspace*{\fill}
\clearpage
\vspace*{\fill}
\begin{table}[H]
\centering
\begin{threeparttable}
\footnotesize
\begin{tabular}{
    >{\centering\arraybackslash}m{3.2cm}   
    >{\centering\arraybackslash}m{3.8cm}   
    l                                      
    c                                      
    c                                      
}
\toprule
\textbf{Topic} & \textbf{Subtopic} & \textbf{Task} & \textbf{Task (\%)} & \textbf{Overall (\%)} \\
\midrule

\multirow{13}{*}{\centering\shortstack{Travel \& Leisure \\ (6.7)}}

  & \multirow{3}{*}{\centering\shortstack{Flights \& Transportation \\ (40.7)}}
      & Search/filter flights \& transportation & 93.4 & 2.55 \\
  &   & Summarize/analyze flights \& transportation & 63.1 & 1.72 \\
  &   & Book flights \& transportation & 9.5 & 0.26 \\[2pt]
\cline{2-5}

  & \multirow{3}{*}{\centering\shortstack{Lodging \\ (27.3)}}
      & Search/filter lodging & 92.9 & 1.70 \\
  &   & Summarize/analyze lodging information & 67.5 & 1.23 \\
  &   & Book lodging & 5.5 & 0.10 \\[2pt]
\cline{2-5}

  & \multirow{3}{*}{\centering\shortstack{Trip Itineraries \\ (24.8)}}
      & Plan trips & 87.4 & 1.45 \\
  &   & Summarize/analyze trips & 48.9 & 0.81 \\
  &   & Search/filter destinations & 48.8 & 0.81 \\
\cline{2-5}

  & \multirow{4}{*}{\centering\shortstack{Restaurants \\ (5.7)}}
      & Search/filter restaurants & 73.8 & 0.28 \\
  &   & Summarize/analyze restaurant information & 54.2 & 0.21 \\
  &   & Book restaurants & 25.0 & 0.09 \\
  &   & Manage bookings & 5.7 & 0.02 \\[2pt]

\bottomrule
\end{tabular}
\caption{Task Distribution for Travel \& Leisure}
\label{tab:use case-travel-task}
\begin{tablenotes}
\scriptsize
    \item Note: The table shows all tasks under Travel \& Leisure with a share of more than 5\% within the subtopic. The share of the topic among all agentic queries and the share of the subtopic within a topic are shown in parentheses. Task percentage ($P(\text{Task | Topic, Subtopic})$) is the task share within the subtopic. The overall percentage is the task share among all agentic queries. $P(\text{Task Overall}) = P(\text{Topic, Subtopic, Task}) = P(\text{Topic}) \times P(\text{Subtopic | Topic}) \times P(\text{Task | Topic, Subtopic}) $). Note that because task percentage measures the fraction of queries in which a task is present and a query might contain multiple tasks, the task percentages under each subtopic do not sum to 100.
\end{tablenotes}
\end{threeparttable}
\end{table}
\vspace*{\fill}

\begin{table}[htbp]
\centering
\begin{threeparttable}
\tiny
\begin{tabular}{|>{\centering\arraybackslash}m{4cm}|l|c|}
\hline
\textbf{Cluster} & \textbf{Task} & \textbf{Task (\%)} \\
\hline

\multirow{5}{*}{\centering Digital Technology}
  & Assist exercises & 9.1 \\
  & Search/filter products & 6.4 \\
  & Create/edit documents/forms & 6.2 \\
  & Summarize/analyze research information & 6.2 \\
  & Manage settings/profiles & 5.7 \\
\hline

\multirow{5}{*}{\centering Student}
  & Assist exercises & 26.5 \\
  & Summarize/analyze course materials & 10.7 \\
  & Navigate courses & 7.8 \\
  & Create/edit documents/forms & 7.1 \\
  & Summarize/analyze research information & 5.4 \\
\hline

\multirow{5}{*}{\centering Management \& Entrepreneurship}
  & Summarize/analyze research information & 7.7 \\
  & Create/edit documents/forms & 7.5 \\
  & Search/filter research information & 6.8 \\
  & Search/filter products & 6.3 \\
  & Create/edit spreadsheets/data & 5.2 \\
\hline

\multirow{5}{*}{\centering Marketing \& Sales}
  & Create social media posts/messages & 8.1 \\
  & Search/filter social media posts/messages & 8.1 \\
  & Summarize/analyze product information & 8.1 \\
  & Summarize/analyze research information & 8.0 \\
  & Search/filter products & 7.6 \\
\hline

\multirow{5}{*}{\centering Financial Services}
  & Summarize/analyze investment information & 11.9 \\
  & Summarize/analyze research information & 7.0 \\
  & Search/filter products & 6.7 \\
  & Search/filter research information & 6.1 \\
  & Summarize/analyze product information & 5.2 \\
\hline

\multirow{5}{*}{\centering Education}
  & Assist exercises & 19.6 \\
  & Create/edit documents/forms & 9.8 \\
  & Summarize/analyze course materials & 8.6 \\
  & Summarize/analyze research information & 7.0 \\
  & Search/filter research information & 6.9 \\
\hline

\multirow{5}{*}{\centering Arts, Entertainment, \& Design}
  & Search/filter products & 7.8 \\
  & Create/edit documents/forms & 7.4 \\
  & Create/edit multimedia & 7.3 \\
  & Summarize/analyze product information & 6.1 \\
  & Summarize/analyze research information & 5.8 \\
\hline

\multirow{5}{*}{\centering Healthcare \& Human Services}
  & Summarize/analyze research information & 9.5 \\
  & Search/filter research information & 9.4 \\
  & Create/edit documents/forms & 7.9 \\
  & Search/filter products & 6.6 \\
  & Assist exercises & 5.5 \\
\hline

\multirow{5}{*}{\centering Advanced Manufacturing}
  & Search/filter products & 20.9 \\
  & Summarize/analyze product information & 18.1 \\
  & Summarize/analyze research information & 7.8 \\
  & Search/filter research information & 7.5 \\
  & Assist exercises & 6.6 \\
\hline

\multirow{5}{*}{\centering Public Service \& Safety}
  & Search/filter research information & 15.7 \\
  & Summarize/analyze research information & 15.1 \\
  & Create/edit documents/forms & 10.2 \\
  & Search/filter products & 5.1 \\
  & Assist exercises & 4.3 \\
\hline

\multirow{5}{*}{\centering Hospitality, Events, \& Tourism}
  & Search/filter flights \& transportation & 11.9 \\
  & Search/filter lodging & 9.6 \\
  & Search/filter products & 8.3 \\
  & Summarize/analyze flights \& transportation information & 8.1 \\
  & Summarize/analyze lodging information & 7.6 \\
\hline

\multirow{5}{*}{\centering Supply Chain \& Transportation}
  & Search/filter products & 12.7 \\
  & Summarize/analyze product information & 10.6 \\
  & Create/edit documents/forms & 7.4 \\
  & Summarize/analyze research information & 7.2 \\
  & Search/filter research information & 6.4 \\
\hline

\multirow{5}{*}{\centering Construction}
  & Search/filter products & 11.0 \\
  & Summarize/analyze product information & 9.1 \\
  & Summarize/analyze research information & 8.8 \\
  & Search/filter research information & 8.3 \\
  & Create/edit documents/forms & 5.6 \\
\hline

\multirow{5}{*}{\centering Energy \& Natural Resources}
  & Create/edit documents/forms & 10.5 \\
  & Summarize/analyze research information & 10.4 \\
  & Search/filter research information & 10.4 \\
  & Search/filter products & 7.7 \\
  & Summarize/analyze product information & 6.0 \\
\hline

\multirow{5}{*}{\centering Agriculture}
  & Create/edit documents/forms & 10.3 \\
  & Search/filter products & 8.5 \\
  & Summarize/analyze research information & 8.4 \\
  & Search/filter research information & 8.4 \\
  & Summarize/analyze product information & 7.2 \\
\hline

\end{tabular}
\caption{The Top 5 Tasks by Occupation Cluster}
\label{tab:use case-task-occupation}
\begin{tablenotes}
    \item Note: The table shows the top 5 tasks under each occupation cluster. Task percentage ($P(\text{Task | Cluster}) $) is the task share among all agentic queries within a cluster. 
\end{tablenotes}

\end{threeparttable}
\end{table}

\clearpage
\subsubsection{Environments}


\vspace*{\fill}
\begin{table}[H]
\centering
\begin{threeparttable}
\footnotesize
\begin{tabular}{
    >{\centering\arraybackslash}m{3.2cm}   
    >{\centering\arraybackslash}m{3.8cm}   
    l                                      
    c                                      
    c                                      
}
\toprule
\textbf{Topic} & \textbf{Subtopic} & \textbf{Environment} & \textbf{Environment (\%)} & \textbf{Overall (\%)} \\
\midrule

\multirow{45}{*}{\centering\shortstack{Productivity \& Workflow \\ (36.2)}}

  & \multirow{5}{*}{\centering\shortstack{Document \& Form Editing \\ (21.5)}}
      & docs.google.com & 66.6 & 5.18 \\
  &   & notion.so & 6.4 & 0.50 \\
  &   & canva.com & 2.5 & 0.19 \\
  &   & overleaf.com & 2.0 & 0.16 \\
  &   & perplexity.ai & 1.3 & 0.10 \\
\cline{2-5}
  & \multirow{5}{*}{\centering\shortstack{Account Management \\ (20.5)}}
      & perplexity.ai & 10.4 & 0.77 \\
  &   & docs.google.com & 7.6 & 0.56 \\
  &   & settings & 4.0 & 0.30 \\
  &   & github.com & 3.2 & 0.24 \\
  &   & linkedin.com & 3.1 & 0.23 \\
\cline{2-5}
  & \multirow{5}{*}{\centering\shortstack{Email Management \\ (15.8)}}
      & mail.google.com & 69.9 & 4.00 \\
  &   & outlook.office.com & 10.8 & 0.62 \\
  &   & outlook.live.com & 2.9 & 0.17 \\
  &   & mail.yahoo.com & 1.5 & 0.09 \\
  &   & mail.yandex.ru & 0.5 & 0.03 \\[2pt]
\cline{2-5}
  & \multirow{5}{*}{\centering\shortstack{Spreadsheet \& Data Editing \\ (11.1)}}
      & docs.google.com & 78.9 & 3.17 \\
  &   & notion.so & 4.4 & 0.18 \\
  &   & airtable.com & 2.3 & 0.09 \\
  &   & excel.cloud.microsoft & 1.4 & 0.06 \\
  &   & app.powerbi.com & 1.1 & 0.04 \\[2pt]
\cline{2-5}
  & \multirow{5}{*}{\centering\shortstack{Computer Programming \\ (10.3)}}
      & github.com & 30.7 & 1.14 \\
  &   & colab.research.google.com & 5.3 & 0.20 \\
  &   & leetcode.com & 4.9 & 0.18 \\
  &   & aistudio.google.com & 4.2 & 0.16 \\
  &   & script.google.com & 3.1 & 0.12 \\[2pt]
\cline{2-5}
  & \multirow{5}{*}{\centering\shortstack{Investments \& Banking \\ (6.2)}}
      & tradingview.com & 47.3 & 1.06 \\
  &   & binance.com & 5.7 & 0.13 \\
  &   & kite.zerodha.com & 4.9 & 0.11 \\
  &   & groww.in & 4.6 & 0.10 \\
  &   & perplexity.ai & 4.4 & 0.10 \\[2pt]
\cline{2-5}
  & \multirow{5}{*}{\centering\shortstack{Multimedia Editing \\ (6.1)}}
      & canva.com & 42.9 & 0.95 \\
  &   & figma.com & 8.6 & 0.19 \\
  &   & docs.google.com & 5.3 & 0.12 \\
  &   & youtube.com & 5.3 & 0.12 \\
  &   & aistudio.google.com & 3.5 & 0.08 \\[2pt]
\cline{2-5}
  & \multirow{5}{*}{\centering\shortstack{Project Management \\ (5.1)}}
      & app.clickup.com & 9.6 & 0.18 \\
  &   & trello.com & 8.1 & 0.15 \\
  &   & notion.so & 7.3 & 0.13 \\
  &   & linear.app & 6.5 & 0.12 \\
  &   & adsmanager.facebook.com & 5.4 & 0.10 \\[2pt]
\cline{2-5}
  & \multirow{5}{*}{\centering\shortstack{Calendar Management \\ (2.5)}}
      & calendar.google.com & 50.3 & 0.45 \\
  &   & outlook.office.com & 7.7 & 0.07 \\
  &   & meet.google.com & 3.7 & 0.03 \\
  &   & mail.google.com & 3.5 & 0.03 \\
  &   & teams.microsoft.com & 1.4 & 0.01 \\

\bottomrule
\end{tabular}
\caption{The Top 5 Environments Distribution for Productivity \& Workflow}
\label{tab:use case-productivity-env}
\begin{tablenotes}
\scriptsize
    \item Note: The table shows the top 5 environments under Productivity \& Workflow. The share of the topic among all agentic queries and the share of the subtopic within a topic are shown in parentheses. Environment percentage ($P(\text{Environment | Topic, Subtopic})$) is the environment share within the subtopic. The overall percentage is the environment share among all agentic queries. $P(\text{Topic, Subtopic, Environment}) = P(\text{Topic}) \times P(\text{Subtopic | Topic}) \times P(\text{Environment | Topic, Subtopic})$. Note that, unlike tasks, an environment is not unique to a subtopic, so $P(\text{Topic, Subtopic, Environment})$ is the share of an environment when it is used under that subtopic and does not equal $P(\text{Environment})$, which is the share under all subtopics.  
\end{tablenotes}
\end{threeparttable}
\end{table}
\vspace*{\fill}

\clearpage
\vspace*{\fill} 
\begin{table}[H]
\centering
\begin{threeparttable}
\footnotesize
\begin{tabular}{
    >{\centering\arraybackslash}m{3.2cm}   
    >{\centering\arraybackslash}m{3.6cm}   
    l                                      
    c                                      
    c                                      
}
\toprule
\textbf{Topic} & \textbf{Subtopic} & \textbf{Environment} & \textbf{Environment (\%)} & \textbf{Overall (\%)} \\
\midrule

\multirow{11}{*}{\centering\shortstack{Learning \& Research \\ (20.8)}}

  & \multirow{5}{*}{\centering\shortstack{Courses \\ (61.9)}}
      & coursera.org              & 18.0 & 2.32 \\
  &   & netacad.com              & 15.6 & 2.01 \\
  &   & canvas.com               & 12.6 & 1.62 \\
  &   & learning.mheducation.com & 8.3  & 1.07 \\
  &   & docs.google.com     & 6.4  & 0.82 \\
  \cline{2-5}

  & \multirow{6}{*}{\centering\shortstack{Research \\ (37.9)}}
      & youtube.com          & 17.8 & 1.40 \\
  &   & perplexity.ai        & 6.1  & 0.48 \\
  &   & github.com           & 5.8  & 0.46 \\
  &   & maps.google.com      & 5.4  & 0.43 \\
  &   & docs.google.com & 3.9  & 0.31 \\
  &   &                      &      &      \\[-6pt]

\bottomrule
\end{tabular}
\caption{The Top 5 Environments Distribution for Learning \& Research}
\label{tab:use case-learning-env}
\begin{tablenotes}
\scriptsize

    \item Note: The table shows the top 5 environments under Learning \& Research. The share of topics and subtopics among all agentic queries is shown in parentheses. Environment percentage ($P(\text{Environment | Topic, Subtopic})$) is the environment share within the subtopic. The overall percentage is the environment share among all agentic queries. $P(\text{Topic, Subtopic, Environment}) = P(\text{Topic}) \times P(\text{Subtopic | Topic}) \times P(\text{Environment | Topic, Subtopic})$. Note that, unlike tasks, an environment is not unique to a subtopic, so $P(\text{Topic, Subtopic, Environment})$ is the share of an environment when it is used under that subtopic and does not equal $P(\text{Environment})$, which is the share under all subtopics.  
\end{tablenotes}
\end{threeparttable}
\end{table}
\vspace*{\fill}

\clearpage
\vspace*{\fill}
\begin{table}[H]
\centering
\begin{threeparttable}
\footnotesize
\begin{tabular}{
    >{\centering\arraybackslash}m{3.2cm}   
    >{\centering\arraybackslash}m{3.6cm}   
    l                                      
    c                                      
    c                                      
}
\toprule
\textbf{Topic} & \textbf{Subtopic} & \textbf{Environment} & \textbf{Environment (\%)} & \textbf{Overall (\%)} \\
\midrule

\multirow{36}{*}{\centering\shortstack{Media \& Entertainment \\ (15.8)}}

  & \multirow{5}{*}{\centering\shortstack{Social Media \& Messaging \\ (42.4)}}
      & instagram.com  & 21.3 & 1.43 \\
  &   & x.com          & 18.0 & 1.21 \\
  &   & whatsapp.com   & 13.6 & 0.91 \\
  &   & facebook.com   & 10.1 & 0.68 \\
  &   & linkedin       &  6.1 & 0.41 \\
  \cline{2-5}

  & \multirow{5}{*}{\centering\shortstack{Movies, TV, \& Videos \\ (20.1)}}
      & youtube.com         & 89.9 & 2.85 \\
  &   & netflix.com        & 4.1  & 0.13 \\
  &   & in.bookmyshow.com  & 1.3  & 0.04 \\
  &   & twitch.tv          & 0.7  & 0.02 \\
  &   & tiktok.com         & 0.6  & 0.02 \\
  \cline{2-5}

  & \multirow{5}{*}{\centering\shortstack{Online Games \\ (19.6)}}
      & chess.com              & 32.5 & 1.01 \\
  &   & store.steampowered.com & 15.0 & 0.46 \\
  &   & nytimes.com            & 14.2 & 0.44 \\
  &   & roblox.com             &  6.8 & 0.21 \\
  &   & humanbenchmark.com     &  5.4 & 0.17 \\
  \cline{2-5}

  & \multirow{5}{*}{\centering\shortstack{Music \& Podcasts \\ (10.7)}}
      & open.spotify.com   & 46.1 & 0.78 \\
  &   & youtube.com        & 39.9 & 0.67 \\
  &   & suno.com           &  6.2 & 0.10 \\
  &   & soundcloud.com     &  2.7 & 0.05 \\
  &   & music.apple.com    &  2.4 & 0.04 \\
  \cline{2-5}

  & \multirow{5}{*}{\centering\shortstack{News \\ (3.8)}}
      & youtube.com        & 21.0 & 0.13 \\
  &   & trends.google.com  & 13.6 & 0.08 \\
  &   & nytimes.com        &  8.7 & 0.05 \\
  &   & perplexity.ai      &  7.9 & 0.05 \\
  &   & x.com              &  5.8 & 0.03 \\
  \cline{2-5}

  & \multirow{6}{*}{\centering\shortstack{Sports \\ (2.7)}}
      & youtube.com        & 28.3 & 0.12 \\
  &   & fantasy.espn.com  & 20.1 & 0.09 \\
  &   & sleeper.com       &  7.8 & 0.03 \\
  &   & sofascore.com     &  3.3 & 0.01 \\
  &   & livescore.in      &  1.8 & 0.01 \\
  &   &                   &      &     \\[-6pt]

\bottomrule
\end{tabular}
\caption{The Top 5 Environments Distribution for Media \& Entertainment}
\label{tab:use case-media-env}
\begin{tablenotes}
\scriptsize
    \item Note: The table shows the top 5 environments under Media \& Entertainment. The share of topics and subtopics among all agentic queries is shown in parentheses. Environment percentage ($P(\text{Environment | Topic, Subtopic})$) is the environment share within the subtopic. The overall percentage is the environment share among all agentic queries. $P(\text{Topic, Subtopic, Environment}) = P(\text{Topic}) \times P(\text{Subtopic | Topic}) \times P(\text{Environment | Topic, Subtopic})$. Note that, unlike tasks, an environment is not unique to a subtopic, so $P(\text{Topic, Subtopic, Environment})$ is the share of an environment when it is used under that subtopic and does not equal $P(\text{Environment})$, which is the share under all subtopics.  
\end{tablenotes}
\end{threeparttable}
\end{table}
\vspace*{\fill}

\clearpage
\vspace*{\fill}
\begin{table}[H]
\footnotesize
\centering
\begin{threeparttable}
\begin{tabular}{
    >{\centering\arraybackslash}m{3.2cm}   
    >{\centering\arraybackslash}m{3.6cm}   
    l                                      
    c                                      
    c                                      
}
\toprule
\textbf{Topic} & \textbf{Subtopic} & \textbf{Environment} & \textbf{Environment (\%)} & \textbf{Overall (\%)} \\
\midrule

\multirow{11}{*}{\centering\shortstack{Shopping \& Commerce \\ (10.0)}}

  & \multirow{5}{*}{\centering\shortstack{Goods \\ (89.0)}}
      & amazon.com            & 43.2 & 3.84 \\
  &   & flipkart.com          & 6.2  & 0.55 \\
  &   & admin.shopify.com     & 5.3  & 0.47 \\
  &   & alibaba.com           & 3.7  & 0.33 \\
  &   & ozon.ru               & 3.4  & 0.30 \\
  \cline{2-5}

  & \multirow{6}{*}{\centering\shortstack{Services \\ (10.3)}}
      & perplexity.ai         & 12.2 & 0.13 \\
  &   & maps.google.com       & 8.7  & 0.09 \\
  &   & amazon.com            & 5.1  & 0.05 \\
  &   & fiverr.com            & 5.1  & 0.05 \\
  &   & avito.ru              & 4.1  & 0.04 \\
  &   &                      &      &     \\[-6pt]

\bottomrule
\end{tabular}
\caption{The Top 5 Environments Distribution for Shopping \& Commerce}
\label{tab:use case-shopping-env}
\begin{tablenotes}
\scriptsize
    \item Note: The table shows the top 5 environments under Shopping \& Commerce. The share of topics and subtopics among all agentic queries is shown in parentheses. Environment percentage ($P(\text{Environment | Topic, Subtopic})$) is the environment share within the subtopic. The overall percentage is the environment share among all agentic queries. $P(\text{Topic, Subtopic, Environment}) = P(\text{Topic}) \times P(\text{Subtopic | Topic}) \times P(\text{Environment | Topic, Subtopic})$. Note that, unlike tasks, an environment is not unique to a subtopic, so $P(\text{Topic, Subtopic, Environment})$ is the share of an environment when it is used under that subtopic and does not equal $P(\text{Environment})$, which is the share under all subtopics.  
\end{tablenotes}

\end{threeparttable}
\end{table}
\vspace*{\fill}

\clearpage
\vspace*{\fill}
\begin{table}[H]
\centering
\begin{threeparttable}
\footnotesize
\begin{tabular}{
    >{\centering\arraybackslash}m{3.2cm}   
    >{\centering\arraybackslash}m{3.6cm}   
    l                                      
    c                                      
    c                                      
}
\toprule
\textbf{Topic} & \textbf{Subtopic} & \textbf{Environment} & \textbf{Environment (\%)} & \textbf{Overall (\%)} \\
\midrule

\multirow{10}{*}{\centering\shortstack{Job \& Career \\ (7.1)}}

  & \multirow{5}{*}{\centering\shortstack{Professional Networking \\ (50.1)}}
      & linkedin.com      & 92.5 & 3.29 \\
  &   & upwork.com        & 1.3  & 0.05 \\
  &   & app.apollo.io     & 0.8  & 0.03 \\
  &   & naukri.com        & 0.6  & 0.02 \\
  &   & instagram.com     & 0.5  & 0.02 \\
  \cline{2-5}

  & \multirow{5}{*}{\centering\shortstack{Job Search \& Application \\ (49.5)}}
      & linkedin.com        & 60.2 & 2.12 \\
  &   & naukri.com          & 6.3  & 0.22 \\
  &   & ziprecruiter.com    & 2.7  & 0.09 \\
  &   & indeed.com          & 2.5  & 0.09 \\
  &   & dice.com            & 2.4  & 0.08 \\

\bottomrule
\end{tabular}
\caption{The Top 5 Environments Distribution for Job \& Career}
\label{tab:use case-job-env}
\begin{tablenotes}
\scriptsize
    \item Note: The table shows the top 5 environments under Job \& Career. The share of topics and subtopics among all agentic queries is shown in parentheses. Environment percentage ($P(\text{Environment | Topic, Subtopic})$) is the environment share within the subtopic. The overall percentage is the environment share among all agentic queries. $P(\text{Topic, Subtopic, Environment}) = P(\text{Topic}) \times P(\text{Subtopic | Topic}) \times P(\text{Environment | Topic, Subtopic})$. Note that, unlike tasks, an environment is not unique to a subtopic, so $P(\text{Topic, Subtopic, Environment})$ is the share of an environment when it is used under that subtopic and does not equal $P(\text{Environment})$, which is the share under all subtopics.  
\end{tablenotes}

\end{threeparttable}
\end{table}
\vspace*{\fill}

\clearpage
\vspace*{\fill}
\begin{table}[H]
\footnotesize
\centering
\begin{threeparttable}
\begin{tabular}{
    >{\centering\arraybackslash}m{3.2cm}   
    >{\centering\arraybackslash}m{3.6cm}   
    l                                      
    c                                      
    c                                      
}
\toprule
\textbf{Topic} & \textbf{Subtopic} & \textbf{Environment} & \textbf{Environment (\%)} & \textbf{Overall (\%)} \\
\midrule

\multirow{21}{*}{\centering\shortstack{Travel \& Leisure \\ (6.7)}}

  & \multirow{5}{*}{\centering\shortstack{Flights \& Transportation \\ (40.7)}}
      & skyscanner.com      & 35.6 & 0.97 \\
  &   & maps.google.com     & 18.4 & 0.50 \\
  &   & makemytrip.com      & 7.2  & 0.20 \\
  &   & irctc.co.in         & 6.1  & 0.17 \\
  &   & expedia.com         & 4.0  & 0.11 \\
  \cline{2-5}

  & \multirow{5}{*}{\centering\shortstack{Lodging \\ (27.3)}}
      & booking.com         & 54.9 & 1.00 \\
  &   & airbnb.com          & 19.5 & 0.36 \\
  &   & expedia.com         & 4.6  & 0.08 \\
  &   & maps.google.com     & 4.0  & 0.07 \\
  &   & agoda.com           & 3.5  & 0.06 \\
  \cline{2-5}

  & \multirow{5}{*}{\centering\shortstack{Trip Itineraries \\ (24.8)}}
      & maps.google.com         & 85.2 & 1.42 \\
  &   & docs.google.com    & 4.2  & 0.07 \\
  &   & yandex.ru               & 1.3  & 0.02 \\
  &   & skyscanner.com          & 1.2  & 0.02 \\
  &   & booking.com             & 0.8  & 0.01 \\
  \cline{2-5}

  & \multirow{5}{*}{\centering\shortstack{Restaurants \\ (5.7)}}
      & maps.google.com     & 54.7 & 0.21 \\
  &   & opentable.com       & 6.5  & 0.02 \\
  &   & map.naver.com       & 5.2  & 0.02 \\
  &   & swiggy.com          & 4.6  & 0.02 \\
  &   & ubereats.com        & 3.9  & 0.01 \\

\bottomrule
\end{tabular}

\caption{The Top 5 Environments Distribution for Travel \& Leisure}
\label{tab:use case-travel-env}
\begin{tablenotes}
\scriptsize
    \item Note: The table shows the top 5 environments under Travel \& Leisure. The share of topics and subtopics among all agentic queries is shown in parentheses. Environment percentage ($P(\text{Environment | Topic, Subtopic})$) is the environment share within the subtopic. The overall percentage is the environment share among all agentic queries. $P(\text{Topic, Subtopic, Environment}) = P(\text{Topic}) \times P(\text{Subtopic | Topic}) \times P(\text{Environment | Topic, Subtopic})$. Note that, unlike tasks, an environment is not unique to a subtopic, so $P(\text{Topic, Subtopic, Environment})$ is the share of an environment when it is used under that subtopic and does not equal $P(\text{Environment})$, which is the share under all subtopics.  
\end{tablenotes}

\end{threeparttable}
\end{table}
\vspace*{\fill}

\clearpage
\vspace*{\fill}
\begin{table}[H]
\centering
\begin{threeparttable}
\footnotesize
\begin{tabular}{
    >{\centering\arraybackslash}m{3.2cm}   
    >{\centering\arraybackslash}m{3.6cm}   
    c                                      
}
\toprule
\textbf{Topic} & \textbf{Subtopic} & \textbf{Sum of the Top 5 Environments (\%)} \\
\midrule

\multirow{9}{*}{\centering\shortstack{Productivity \& Workflow}}
  & \centering\shortstack{Spreadsheet \& Data Editing} & 88.1 \\
  & \centering\shortstack{Email Management} & 85.1 \\
  & \centering\shortstack{Document \& Form Editing} & 78.8 \\
  & \centering\shortstack{Investments \& Banking} & 66.9 \\
  & \centering\shortstack{Calendar Management} & 66.6 \\
  & \centering\shortstack{Multimedia Editing} & 65.6 \\
  & \centering\shortstack{Computer Programming} & 48.2 \\
  & \centering\shortstack{Project Management} & 36.9 \\
  & \centering\shortstack{Account Management} & 28.3 \\
\midrule

\multirow{2}{*}{\centering\shortstack{Learning \& Research}}
  & \centering\shortstack{Courses} & 60.9 \\
  & \centering\shortstack{Research} & 39.0 \\
\midrule

\multirow{6}{*}{\centering\shortstack{Media \& Entertainment}}
  & \centering\shortstack{Music \& Podcasts} & 97.3 \\
  & \centering\shortstack{Movies, TV, \& Videos} & 96.6 \\
  & \centering\shortstack{Online Games} & 73.9 \\
  & \centering\shortstack{Social Media \& Messaging} & 69.1 \\
  & \centering\shortstack{Sports} & 61.3 \\
  & \centering\shortstack{News} & 57.0 \\
\midrule

\multirow{2}{*}{\centering\shortstack{Shopping \& Commerce}}
  & \centering\shortstack{Goods} & 61.8 \\
  & \centering\shortstack{Services} & 35.2 \\
\midrule

\multirow{2}{*}{\centering\shortstack{Job \& Career}}
  & \centering\shortstack{Professional Networking} & 95.7 \\
  & \centering\shortstack{Job Search \& Application} & 74.1 \\
\midrule

\multirow{4}{*}{\centering\shortstack{Travel \& Leisure}}
  & \centering\shortstack{Trip Itineraries} & 92.7 \\
  & \centering\shortstack{Lodging} & 86.5 \\
  & \centering\shortstack{Restaurants} & 74.9 \\
  & \centering\shortstack{Flights \& Transportation} & 71.3 \\

\bottomrule
\end{tabular}

\caption{Sum of the Top 5 Environment Shares by Topic and Subtopic}
\label{tab:use case-env-concentration}
\begin{tablenotes}
\scriptsize
    \item Note: The table shows the sum of the top 5 environments' shares by topic and subtopic. A higher (lower) share indicates agentic queries are more (less) concentrated in a small number of environments. This metric can be interpreted as the agent usage market share of environments among agent adopters on Comet.  
\end{tablenotes}

\end{threeparttable}
\end{table}
\vspace*{\fill}

\begin{table}[htbp]
\centering
\begin{threeparttable}
\tiny
\begin{tabular}{|>{\centering\arraybackslash}m{4cm}|l|c|}
\hline
\textbf{Cluster} & \textbf{Environment} & \textbf{Environment (\%)} \\
\hline

\multirow{5}{*}{\centering Digital Technology}
  & linkedin.com & 6.5 \\
  & email services combined & 5.4 \\
  & docs.google.com & 4.0 \\
  & youtube.com & 3.3 \\
  & amazon.com & 2.4 \\
\hline

\multirow{5}{*}{\centering Student}
  & docs.google.com & 7.8 \\
  & linkedin.com & 6.1 \\
  & email services combined & 4.3 \\
  & canvas.com & 3.4 \\
  & youtube.com & 3.0 \\
\hline

\multirow{5}{*}{\centering Management \& Entrepreneurship}
  & linkedin.com & 10.3 \\
  & email services combined & 8.8 \\
  & docs.google.com & 7.7 \\
  & youtube.com & 2.0 \\
  & amazon.com & 1.8 \\
\hline

\multirow{5}{*}{\centering Marketing \& Sales}
  & linkedin.com & 7.8 \\
  & docs.google.com & 6.8 \\
  & instagram.com & 6.5 \\
  & x.com & 5.2 \\
  & email services combined & 4.5 \\
\hline

\multirow{5}{*}{\centering Financial Services}
  & email services combined & 6.6 \\
  & docs.google.com & 5.0 \\
  & linkedin.com & 4.3 \\
  & youtube.com & 3.6 \\
  & tradingview.com & 2.6 \\
\hline

\multirow{5}{*}{\centering Education}
  & docs.google.com & 9.9 \\
  & email services combined & 6.5 \\
  & youtube.com & 4.4 \\
  & canvas.com & 2.4 \\
  & amazon.com & 2.1 \\
\hline

\multirow{5}{*}{\centering Arts, Entertainment, \& Design}
  & youtube.com & 7.0 \\
  & email services combined & 5.1 \\
  & docs.google.com & 4.9 \\
  & linkedin.com & 3.6 \\
  & instagram.com & 3.2 \\
\hline

\multirow{5}{*}{\centering Healthcare \& Human Services}
  & email services combined & 7.3 \\
  & docs.google.com & 5.8 \\
  & linkedin.com & 4.7 \\
  & youtube.com & 3.0 \\
  & amazon.com & 2.5 \\
\hline

\multirow{5}{*}{\centering Advanced Manufacturing}
  & email services combined & 5.5 \\
  & linkedin.com & 4.6 \\
  & docs.google.com & 4.3 \\
  & youtube.com & 3.7 \\
  & amazon.com & 3.4 \\
\hline

\multirow{5}{*}{\centering Public Service \& Safety}
  & email services combined & 6.2 \\
  & docs.google.com & 4.9 \\
  & youtube.com & 4.1 \\
  & trends.google.com & 3.9 \\
  & linkedin.com & 2.8 \\
\hline

\multirow{5}{*}{\centering Hospitality, Events, \& Tourism}
  & email services combined & 6.4 \\
  & maps.google.com & 5.6 \\
  & docs.google.com & 5.4 \\
  & booking.com & 3.6 \\
  & skyscanner.com & 2.9 \\
\hline

\multirow{5}{*}{\centering Supply Chain \& Transportation}
  & email services combined & 8.0 \\
  & docs.google.com & 4.9 \\
  & linkedin.com & 3.7 \\
  & amazon.com & 3.0 \\
  & maps.google.com & 2.5 \\
\hline

\multirow{5}{*}{\centering Construction}
  & email services combined & 8.2 \\
  & linkedin.com & 6.3 \\
  & docs.google.com & 4.3 \\
  & youtube.com & 3.1 \\
  & amazon.com & 2.8 \\
\hline

\multirow{5}{*}{\centering Energy \& Natural Resources}
  & email services combined & 7.4 \\
  & docs.google.com & 6.6 \\
  & linkedin.com & 3.8 \\
  & amazon.com & 3.4 \\
  & youtube.com & 2.9 \\
\hline

\multirow{5}{*}{\centering Agriculture}
  & docs.google.com & 7.4 \\
  & email services combined & 6.9 \\
  & youtube.com & 3.6 \\
  & linkedin.com & 3.6 \\
  & amazon.com & 2.4 \\
\hline
\end{tabular}
\caption{The Top 5 Environments by Occupation Cluster}
\label{tab:use case-env-occupation}
\begin{tablenotes}
\tiny
    \item Note: The table shows the top 5 environments by occupation cluster. Environment percentage ($P(\text{Environment | Cluster}) $) is the environment share among all agentic queries in that cluster. docs.google.com includes Google Docs, Sheets, Slides, and Forms. All email accounts are grouped into ``email services combined''. 
\end{tablenotes}
\end{threeparttable}
\end{table}

\clearpage
\subsubsection{Usage Context}

\vspace*{\fill}
\begin{table}[htbp]
\centering
\begin{threeparttable}
\footnotesize
\begin{tabular}{lllr}
\toprule
\textbf{Context} & \textbf{Topic} & \textbf{Subtopic} & \textbf{Subtopic (\%)} \\
\midrule
\multirow{5}{*}{Personal} & 
Shopping \& Commerce & Goods & 15.6 \\
& Media \& Entertainment & Social Media \& Messaging & 9.9 \\
& Productivity \& Workflow & Account Management & 8.0 \\
& Productivity \& Workflow & Email Management & 7.6 \\
& Media \& Entertainment & Online Games & 6.0 \\
\midrule
\multirow{5}{*}{Professional} & 
Productivity \& Workflow & Document \& Form Editing & 13.3 \\
& Job \& Career & Professional Networking & 12.5 \\
& Job \& Career & Job Search \& Application & 11.0 \\
& Productivity \& Workflow & Account Management & 10.2 \\
& Learning \& Research & Research & 8.9 \\
\midrule
\multirow{5}{*}{Educational} 
& Learning \& Research & Courses & 83.9 \\
& Learning \& Research & Research & 5.3 \\
& Productivity \& Workflow & Document \& Form Editing & 5.0 \\
& Productivity \& Workflow & Account Management & 1.1 \\
& Productivity \& Workflow & Computer Programming & 0.9 \\
\bottomrule
\end{tabular}
\caption{The Top 5 Subtopic Distribution by Usage Context}
\label{tab:use case-context-subtopic}
\begin{tablenotes}
\scriptsize
\item Note: The table shows the distribution of the top 5 subtopics by usage context. Subtopic percentage ($P(\text{Subtopic | Context})$) is the subtopic share among all agentic queries in a given usage context.
\end{tablenotes}
\end{threeparttable}
\end{table}
\vspace*{\fill}

\clearpage
\vspace*{\fill}
\begin{table}[H]
\rotatebox{90}{
\begin{threeparttable}
\footnotesize
\begin{tabular}{llllr}
\toprule
\textbf{Context} & \textbf{Topic} & \textbf{Subtopic} & \textbf{Task} & \textbf{Task (\%)} \\
\midrule
\multirow{5}{*}{Personal} & Shopping \& Commerce & Goods & Search/filter products & 8.6 \\
& Shopping \& Commerce & Goods & Summarize/analyze product information & 6.5 \\
& Travel \& Leisure & Flights \& Transportation & Search/filter flights \& transportation & 3.4 \\
& Media \& Entertainment & Social Media \& Messaging & Search/filter social media posts/messages & 3.4 \\
& Productivity \& Workflow & Document \& Form Editing & Create/edit documents/forms & 3.2 \\
\midrule
\multirow{5}{*}{Professional} & Productivity \& Workflow & Document \& Form Editing & Create/edit documents/forms & 8.1 \\
& Learning \& Research & Research & Summarize/analyze research information & 5.9 \\
& Job \& Career & Job Search \& Application & Complete applications & 5.4 \\
& Productivity \& Workflow & Account Management & Manage settings/profiles & 5.1 \\
& Job \& Career & Professional Networking & Search/filter professional profiles & 4.8 \\
\midrule
\multirow{5}{*}{Educational} & Learning \& Research & Courses & Assist exercises & 48.1 \\
& Learning \& Research & Courses & Summarize/analyze course materials & 18.8 \\
& Learning \& Research & Courses & Navigate courses & 16.0 \\
& Productivity \& Workflow & Document \& Form Editing & Create/edit documents/forms & 3.6 \\
& Learning \& Research & Research & Summarize/analyze research information & 3.4 \\
\bottomrule
\end{tabular}
\caption{The Top 5 Task Distribution by Usage Context}
\label{tab:use case-context-task}
\begin{tablenotes}
\scriptsize
\item Note: The table shows the distribution of the top 5 tasks by usage context. Task percentage ($P(\text{Task | Context})$) is the task share among all agentic queries in a given usage context.
\end{tablenotes}
\end{threeparttable}
}
\end{table}
\vspace*{\fill}

\clearpage
\vspace*{\fill}
\begin{table}[H]
\centering
\begin{threeparttable}
\small
\begin{tabular}{llr}
\toprule
\textbf{Context} & \textbf{Environment} & \textbf{Environment (\%)} \\
\midrule
\multirow{5}{*}{Personal} & email services combined & 14.5 \\
& youtube.com & 10.8 \\
& docs.google.com & 10.7 \\
& amazon.com & 6.3 \\
& maps.google.com & 3.8 \\
\midrule
\multirow{5}{*}{Professional} & linkedin.com & 29.6 \\
& docs.google.com & 11.4 \\
& email services combined & 9.6 \\
& github.com & 3.8 \\
& admin.shopify.com & 2.8 \\
\midrule
\multirow{5}{*}{Educational} & docs.google.com & 14.8 \\
& coursera.org & 14.6 \\
& netacad.com & 12.7 \\
& canvas.com & 10.2 \\
& learning.mheducation.com & 6.8 \\
\bottomrule
\end{tabular}
\caption{The Top 5 Environment Distribution by Usage Context}
\label{tab:use case-context-env}
\begin{tablenotes}
\footnotesize
\item Note: The table shows the distribution of the top 5 environments by usage context. Environment percentage ($P(\text{Environment | Context})$) is the environment share among all agentic queries in a given usage context. docs.google.com includes Google Docs, Sheets, Slides, and Forms. All email domains are grouped into ``email services combined''.
\end{tablenotes}
\end{threeparttable}
\end{table}
\vspace*{\fill}

\section{Agent Demo}\label{agent demo}
\subsection{Sample Agentic Queries}
\begin{figure}[H]
    \centering
    \includegraphics[width=\linewidth]{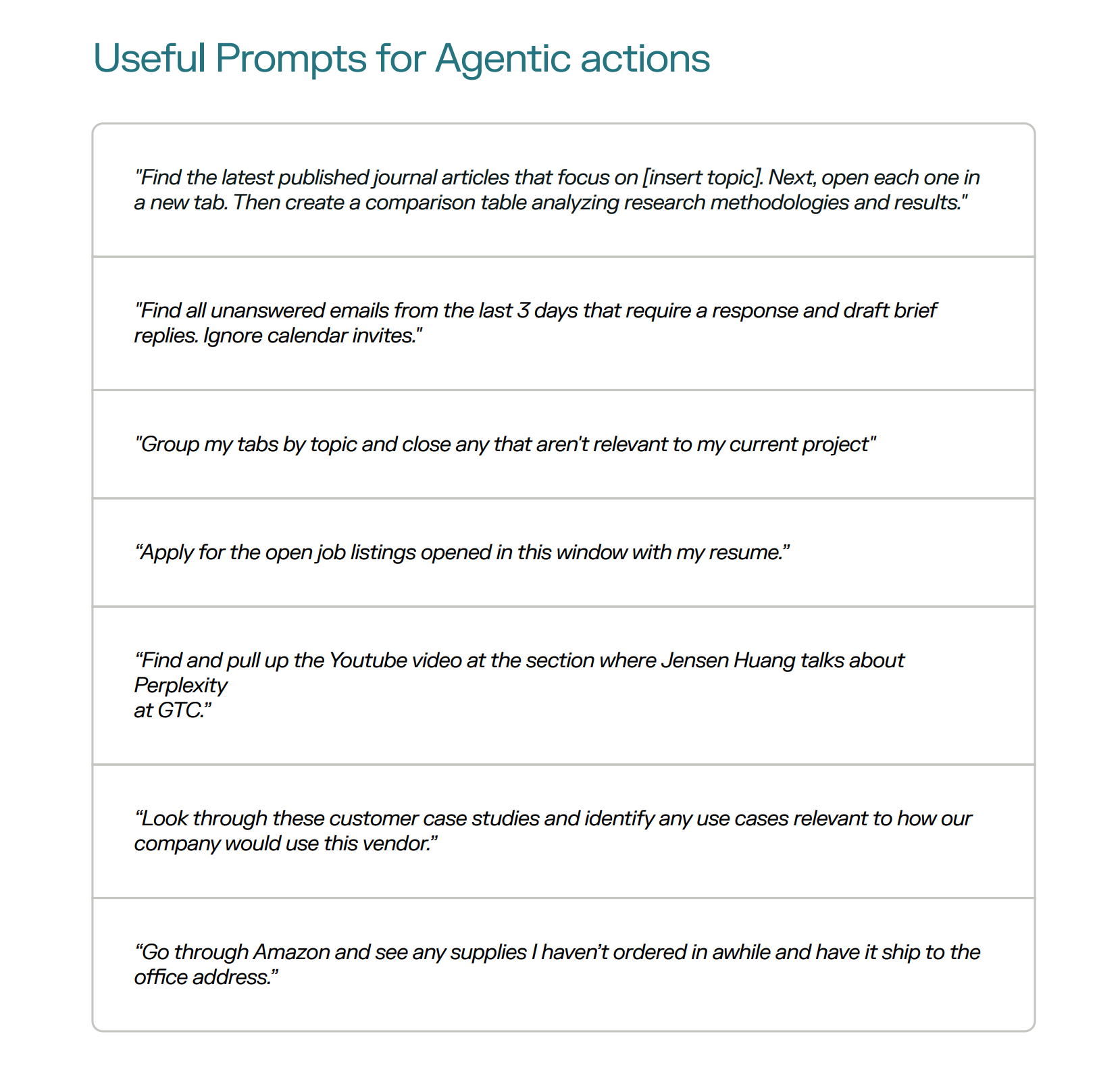}
    \caption{Sample Agentic Queries from Perplexity at Work}
    \label{fig:agent prompt}
\end{figure}

\clearpage
\vspace*{\fill}

\begin{table}[H]
\centering
\rotatebox{90}{
\begin{threeparttable}
\footnotesize
\begin{tabular}{p{0.35\linewidth} p{0.2\linewidth} p{0.2\linewidth} p{0.35\linewidth}}
\toprule
\textbf{Sample Query} & \textbf{Topic} & \textbf{Subtopic} & \textbf{Task} \\
\midrule
Find the latest published journal articles... & Learning \& Research & Research & Search/filter research information, Summarize/analyze research information\\
Find all unanswered emails... & Productivity \& Workflow & Email Management & Search/filter emails, Create/edit emails \\
Group my tabs by topic... & Productivity \& Workflow & Account Management & Manage settings/profiles \\
Apply for the open job listings... & Job \& Career & Job Search \& Application & Complete application \\
Find and pull up the YouTube video... & Media \& Entertainment & Movies, TV, \& Videos & Search/filter videos, Navigate within videos\\
Look through these customer case studies... & Learning \& Research & Research & Search/filter research information, Summarize/analyze research information\\
Go through Amazon... & Shopping \& Commerce & Goods & Search/filter products, Make product purchase\\ 
\bottomrule
\end{tabular}
\begin{tablenotes}
\footnotesize
\item Note: The table shows how the sample agentic queries map to our taxonomy.
\end{tablenotes}
\caption{Sample Agentic Queries Mapped to the Taxonomy}
\label{tab:query-taxonomy}
\end{threeparttable}}
\end{table}
\vspace*{\fill}

\clearpage 
\subsection{Agent Behavior for a Real Task}

The prompt asks the agent to find a round-trip flight between Boston and San Diego to attend the 2025 NeurIPS conference. The user prefers late-night direct flights and a lower price. The Figures below show some key steps in how the agent completes the task.\footnote{See \url{https://www.perplexity.ai/search/search-for-the-best-flights-be-TNxDicWmSiW9gmcpa2pHNg\#0} for all the steps.} 

\vspace*{\fill}
\begin{figure}[H]
    \centering
    \includegraphics[width=\linewidth]{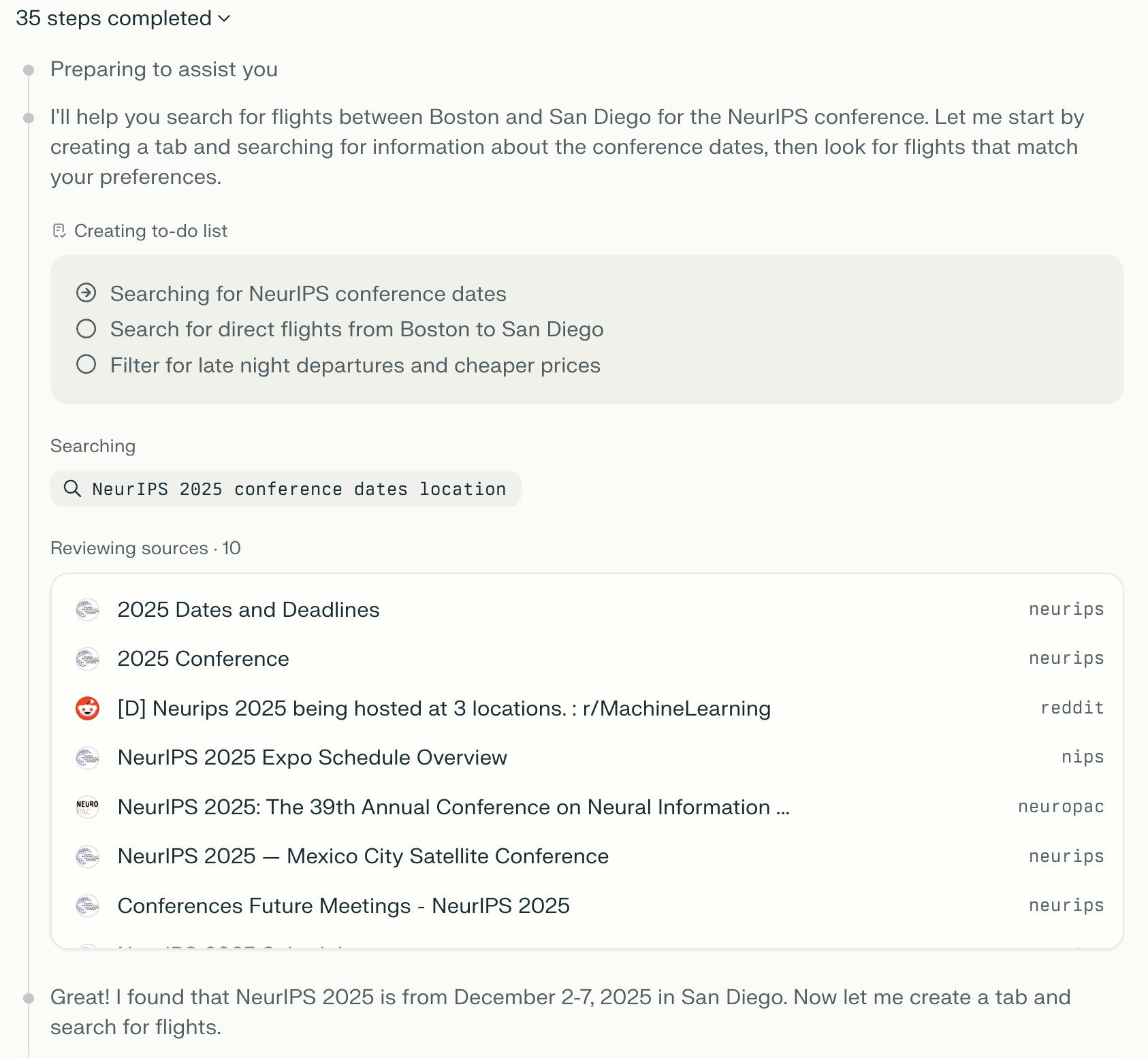}
    \caption{Agent Behavior 1}
    \label{fig:agent flow}
      \begin{minipage}[t]{\textwidth}
      \raggedright
      \footnotesize
       Note: The agent first creates a to-do list that breaks the task into three steps. Because the flight dates are not specified, the agent first searches for the 2025 NeurIPS conference dates and confirms that they are found.
    \end{minipage}
\end{figure}
\vspace*{\fill}

\clearpage
\vspace*{\fill}
\begin{figure}[H]
    \centering
    \includegraphics[width=\linewidth]{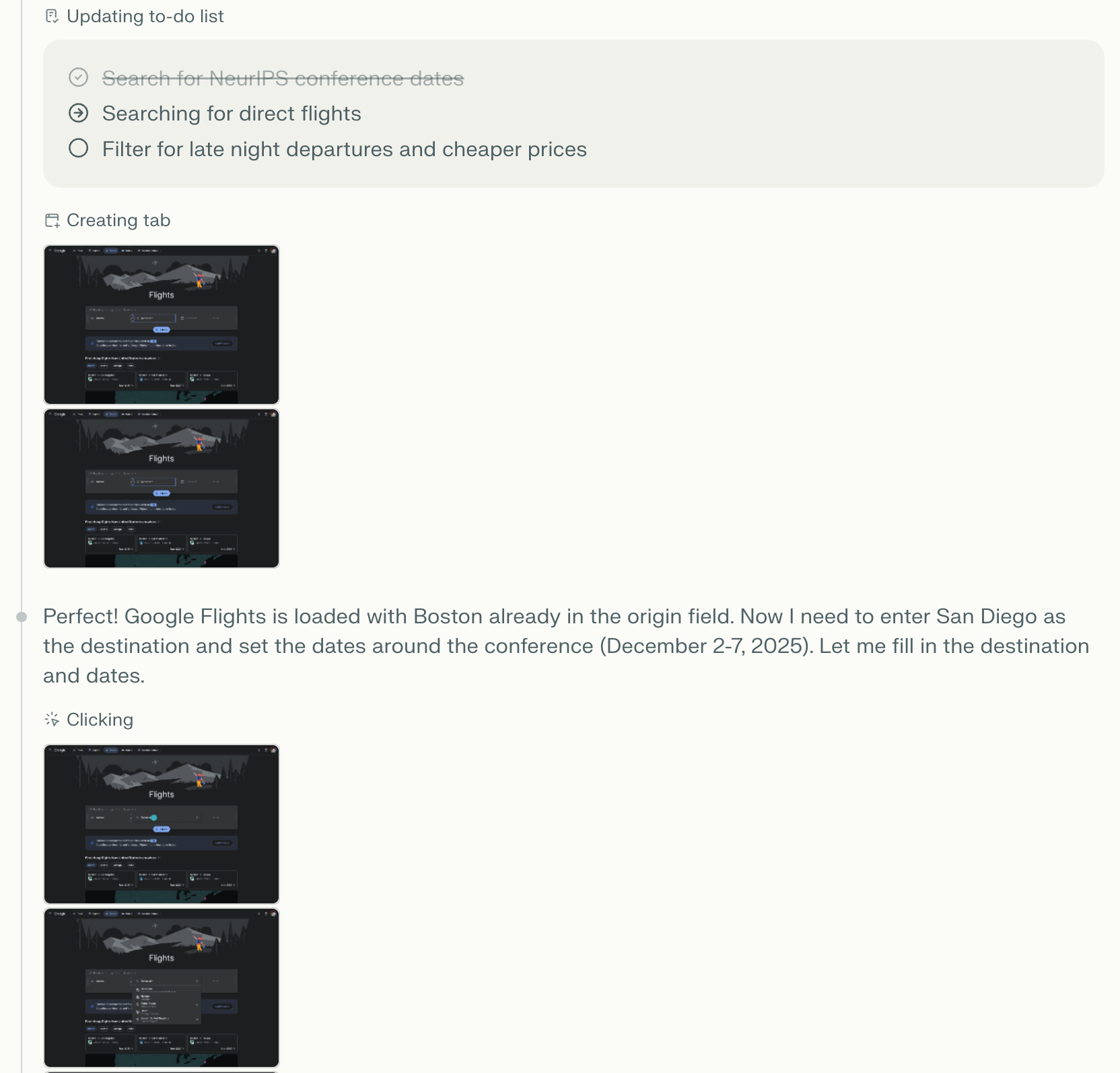}
    \caption{Agent Behavior 2}
    \label{fig:placeholder}
      \begin{minipage}[t]{\textwidth}
      \raggedright
      \footnotesize
       Note: The agent then updates the to-do list and proceeds to the next item by searching flights. It operates on the website by entering destinations and dates.
    \end{minipage}
\end{figure}
\vspace*{\fill}

\clearpage
\vspace*{\fill}
\begin{figure}[H]
    \centering
    \includegraphics[width=\linewidth]{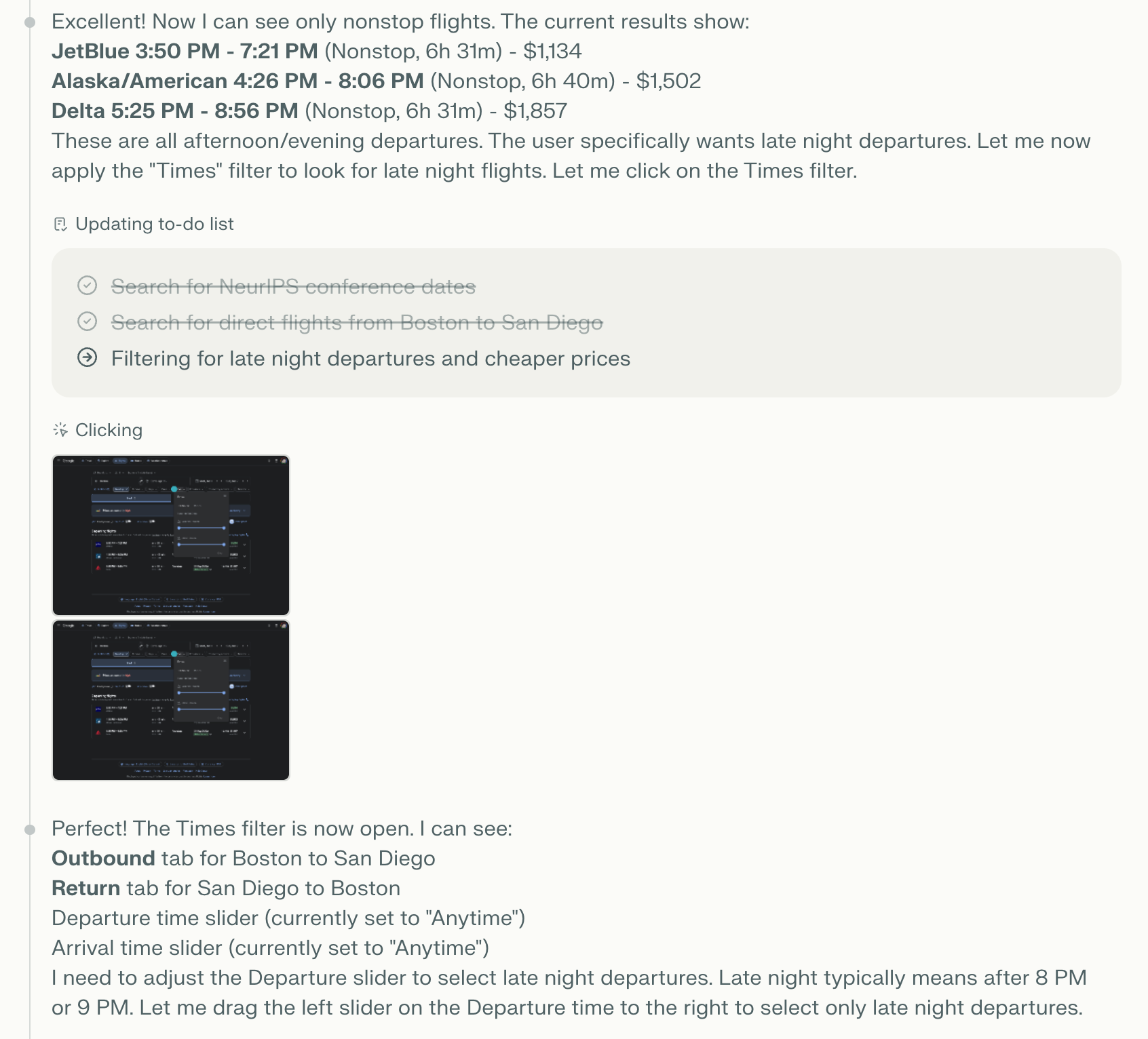}
    \caption{Agent Behavior 3}
    \label{fig:placeholder}
        \begin{minipage}[t]{\textwidth}
      \raggedright
      \footnotesize
       Note: The agent then applies the nonstop filter to narrow the search to direct flights only. But the flights shown on the website do not match the user's preferred flight time, so the agent applies a time filter to find late-night flights.
    \end{minipage}
\end{figure}
\vspace*{\fill}

\clearpage
\vspace*{\fill}
\begin{figure}[H]
    \centering
    \includegraphics[width=\linewidth]{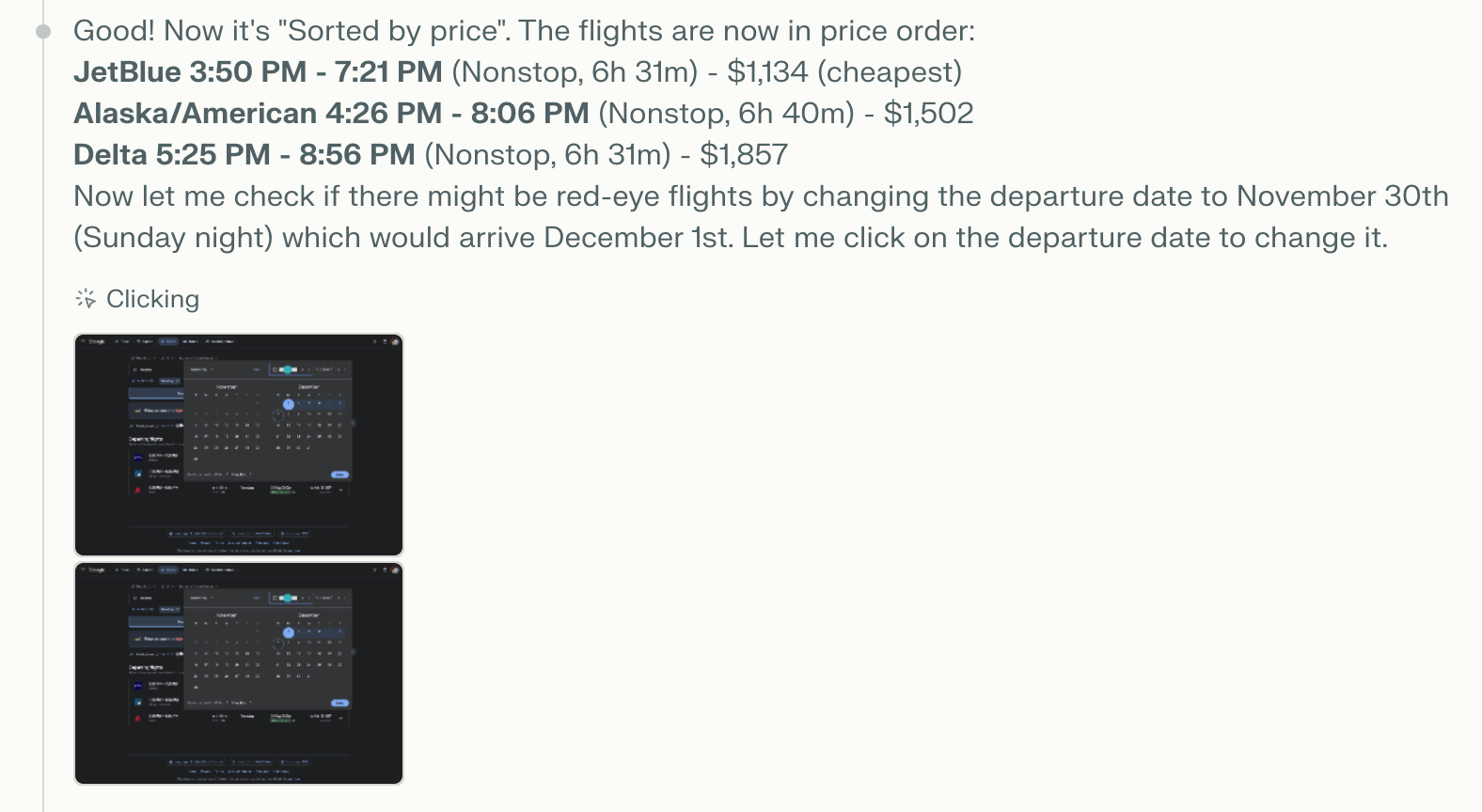}
    \caption{Agent Behavior 4}
    \label{fig:placeholder}
         \begin{minipage}[t]{\textwidth}
      \raggedright
      \footnotesize
       Note: The agent applies a price filter and observes options. Because options are limited, it expands the travel dates to look for red-eye flights.
    \end{minipage}
\end{figure}
\vspace*{\fill}

\clearpage
\vspace*{\fill}
\begin{figure}[H]
    \centering
    \includegraphics[width=\linewidth]{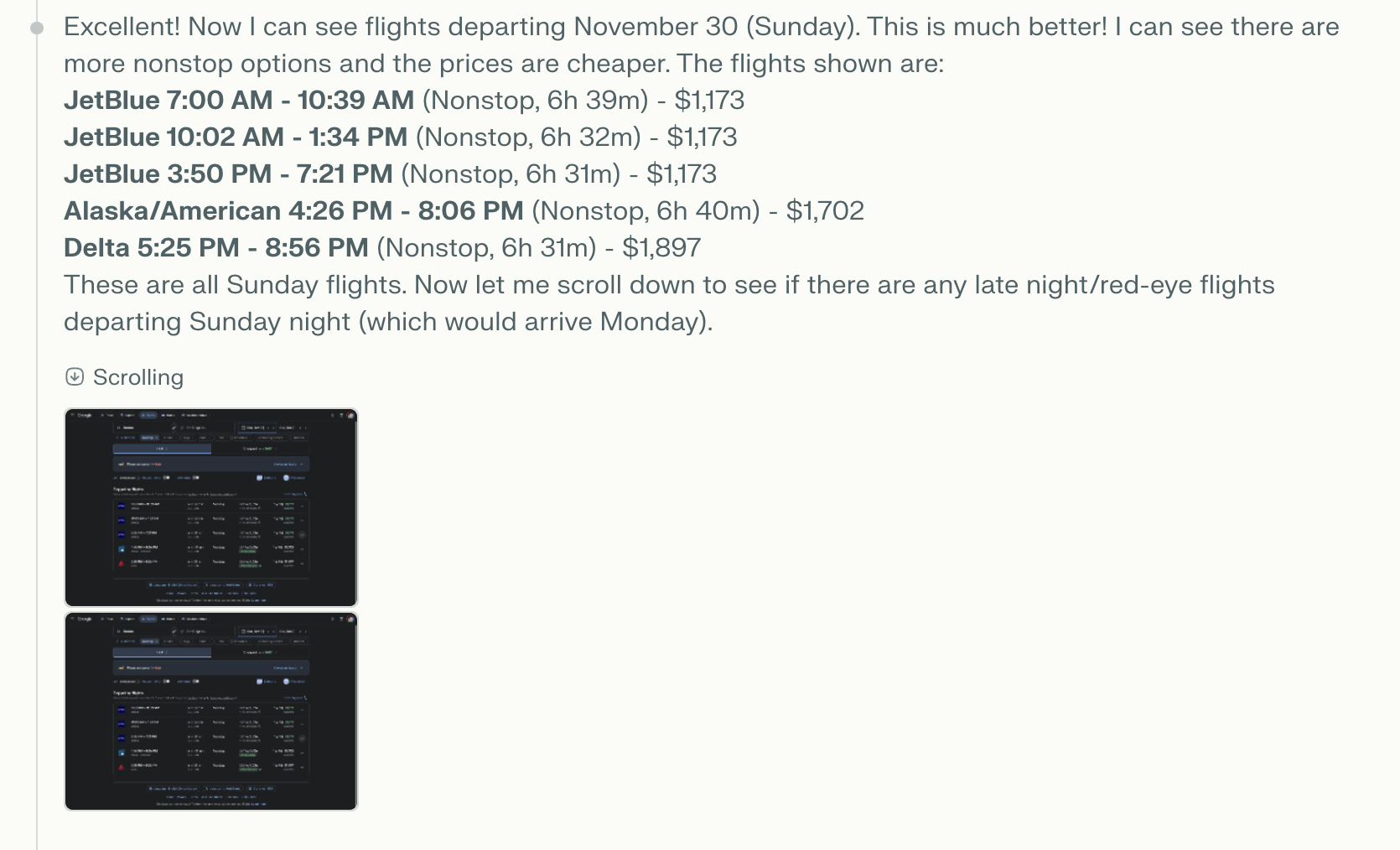}
    \caption{Agent Behavior 5}
    \label{fig:placeholder}
             \begin{minipage}[t]{\textwidth}
      \raggedright
      \footnotesize
       Note: The agent expands the dates and confirms that red-eye flights departing a day earlier are much cheaper.
    \end{minipage}
\end{figure}
\vspace*{\fill}

\clearpage
\vspace*{\fill}
\begin{figure}[H]
    \centering
    \includegraphics[width=\linewidth]{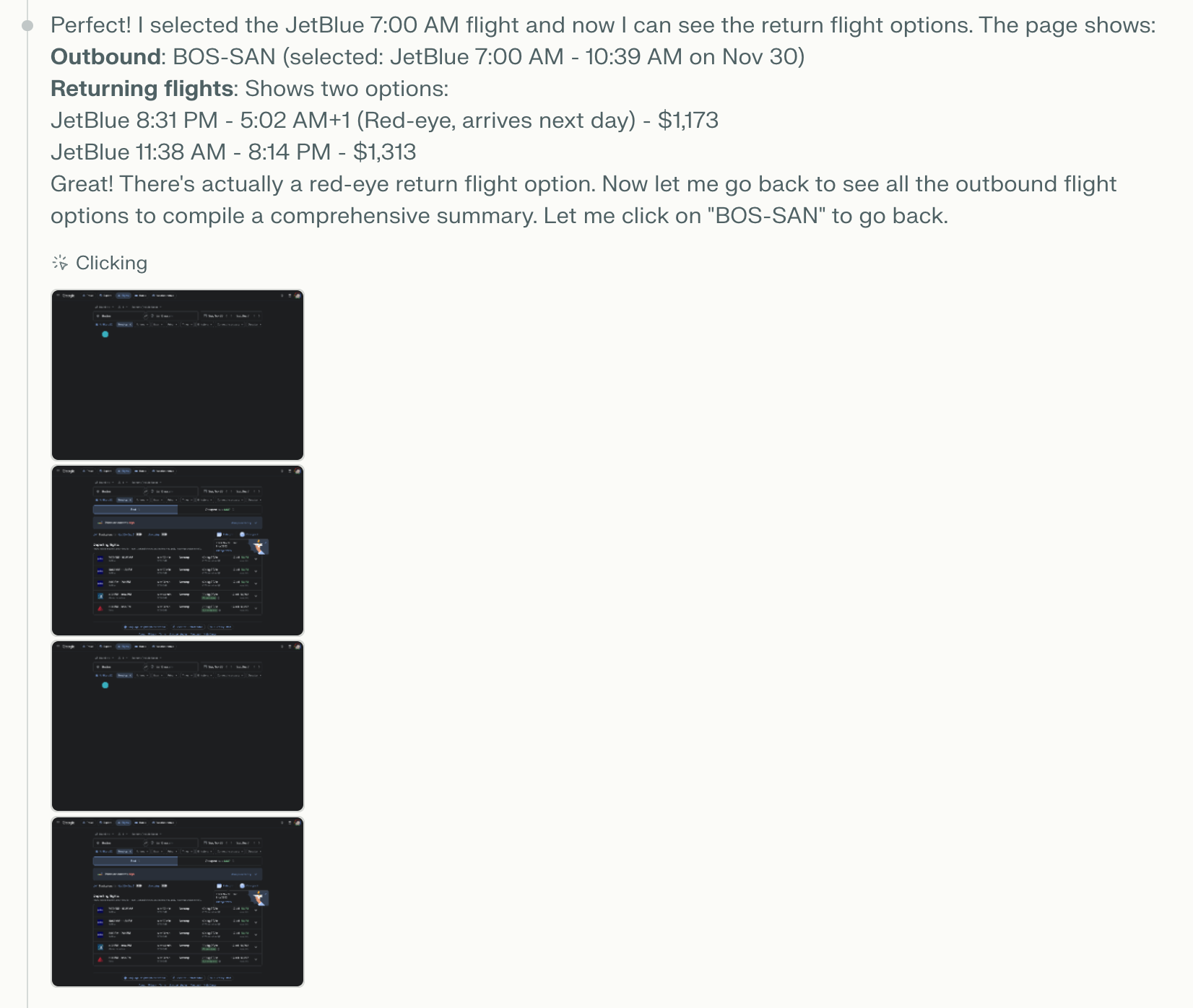}
    \caption{Agent Behavior 6}
    \label{fig:placeholder}
                \begin{minipage}[t]{\textwidth}
      \raggedright
      \footnotesize
       Note: The agent clicks on an outbound flight with the lowest price and proceeds to the inbound flight.
    \end{minipage}
\end{figure}
\vspace*{\fill}

\clearpage
\begin{figure}[h]
    \centering
    \includegraphics[width=.8\linewidth]{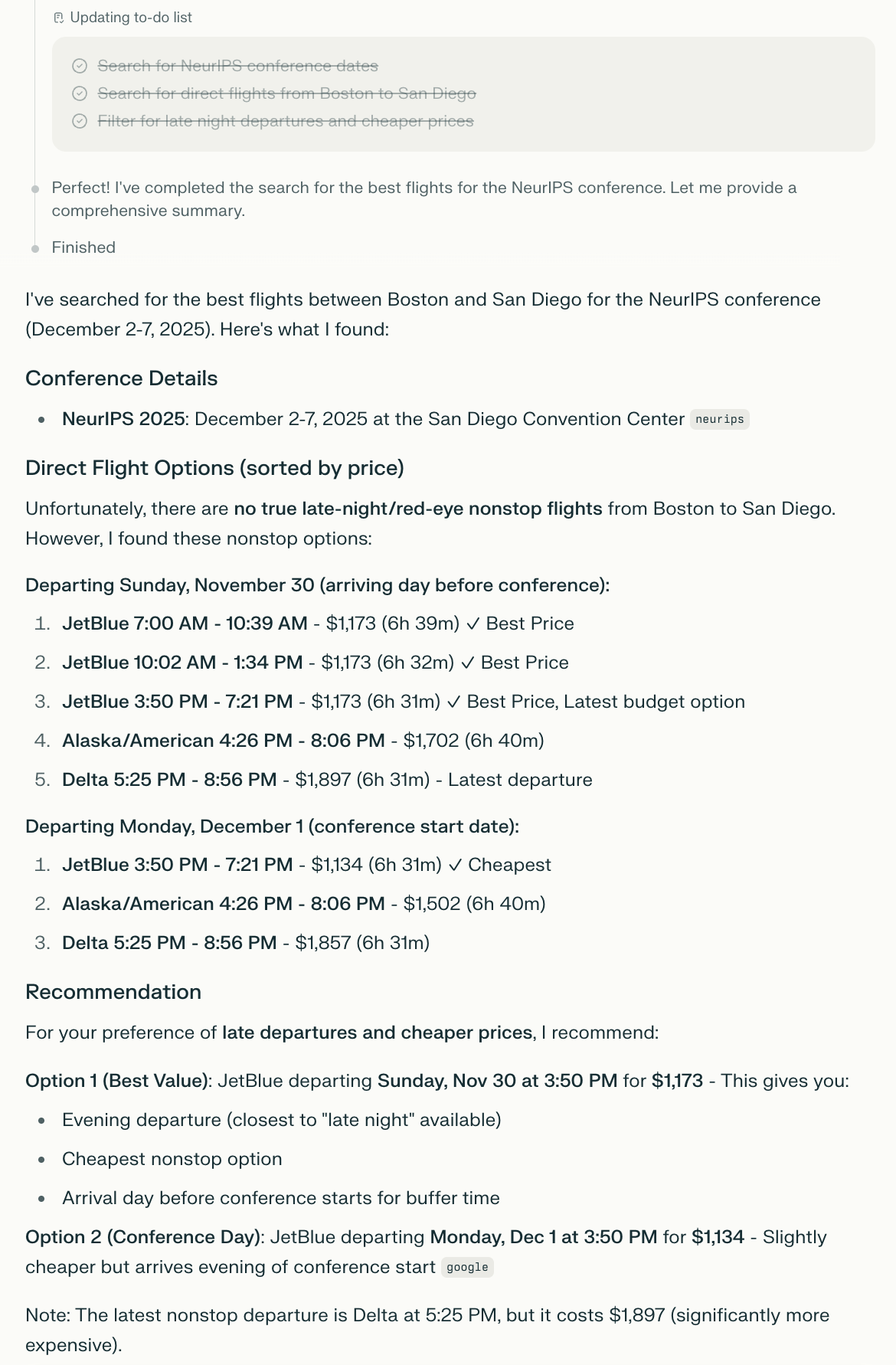}
    \caption{Agent Behavior 7}
    \label{fig:placeholder}
          \begin{minipage}[t]{\textwidth}
      \raggedright
      \footnotesize
       Note: The agent completes the task by presenting the recommendations to the user.
    \end{minipage}
\end{figure}

\clearpage
\section{Early Adopter Survey}\label{survey}
\begin{flushleft}
To better understand who the early users of Comet are, we invited a subset of users to complete a survey in November 2025. A typical respondent is a male aged 35 or older who works full-time in the technology industry, is either a current Perplexity user or has a strong interest in AI-powered browsing, and uses Comet roughly equally across professional and personal contexts. As is often the case with new technology products, we expect user composition to change significantly over time as Comet diffuses into the population.
\end{flushleft}

\section{Validation of Agent Use Cases Classifier}\label{validation}

\begin{flushleft}
We randomly select 1,000 agentic queries for manual labeling.\footnote{Note that because of our focus on agentic queries, we cannot evaluate the classifier's performance on public Q\&A query datasets such as WildChat. \url{https://wildchat.allen.ai/}} These queries are sampled from a larger set previously classified as containing no harmful content. To preserve representativeness, we retain queries that may include personal information, while redacting names, email addresses, physical addresses, and phone numbers. Each query in the sample is independently labeled by two or three annotators using our agentic taxonomy. Out of the 1,000 queries, 370 show disagreement among annotators on what the primary topic and subtopic are. Most disagreements stemmed from variations in labeling quality across annotators.
For each query that shows disagreement, our team manually reviews it and labels it against our taxonomy. The final golden dataset includes the 630 queries on which all annotators agreed, along with the 370 queries we labeled. We validate the classifier against the golden dataset and across multiple runs, and the agreement rates are listed in Table \ref{tab:eval}.
\end{flushleft} 

\begin{table}[htbp]
\centering
\begin{threeparttable}
\small
\caption{Agent Use Case Classification Validation}
\label{tab:eval}
\begin{tabular}{|l|c|c|}
\hline
\textbf{Variable} & \textbf{Golden Dataset (\%)} & \textbf{Across Runs (\%)} \\
\hline
Topic & 89.4 & 97.2 \\
\hline
Subtopic & 83.2 & 94.6\\
\hline
Task & 81.3 & 88.2 \\
\hline
Usage Context  & 82.9 &  96.3\\
\hline
\end{tabular}
\begin{tablenotes}
    \footnotesize
    \item Note: The table shows the agreement rate between the classifier label and the golden dataset and across runs. The tasks are specific to subtopics, so when the classifier disagrees with the golden dataset or across runs on topics or subtopics, the tasks will by definition have zero agreement rates. Therefore, the task agreement rate is conditional on topic- and subtopic-level agreement. The across-runs agreement rate is the average pairwise agreement rate across three runs.
\end{tablenotes}
\end{threeparttable}
\end{table}

\end{document}